%% file: main.tex
\pdfoutput=1
\documentclass[dvipsnames, 11pt]{article}
\usepackage{graphicx} 
\RequirePackage[top=2.25cm, bottom=2.5cm, left=2.5cm, right=2.5cm, columnsep=0.65cm]{geometry}
\usepackage{multirow}
\usepackage{amsmath,amssymb,bm}
\usepackage{booktabs}
\usepackage{xcolor}
\usepackage[numbers]{natbib}
\usepackage{wrapfig}
\usepackage{hyperref}
\usepackage{subcaption}
\usepackage{caption}
\def\method{\textrm{Vidi}} 
\def\ie{\textit{i.e.}}
\def\eg{\textit{e.g.}}
\def\etal{\textit{et al.}}

\title{\includegraphics[width=.8cm]{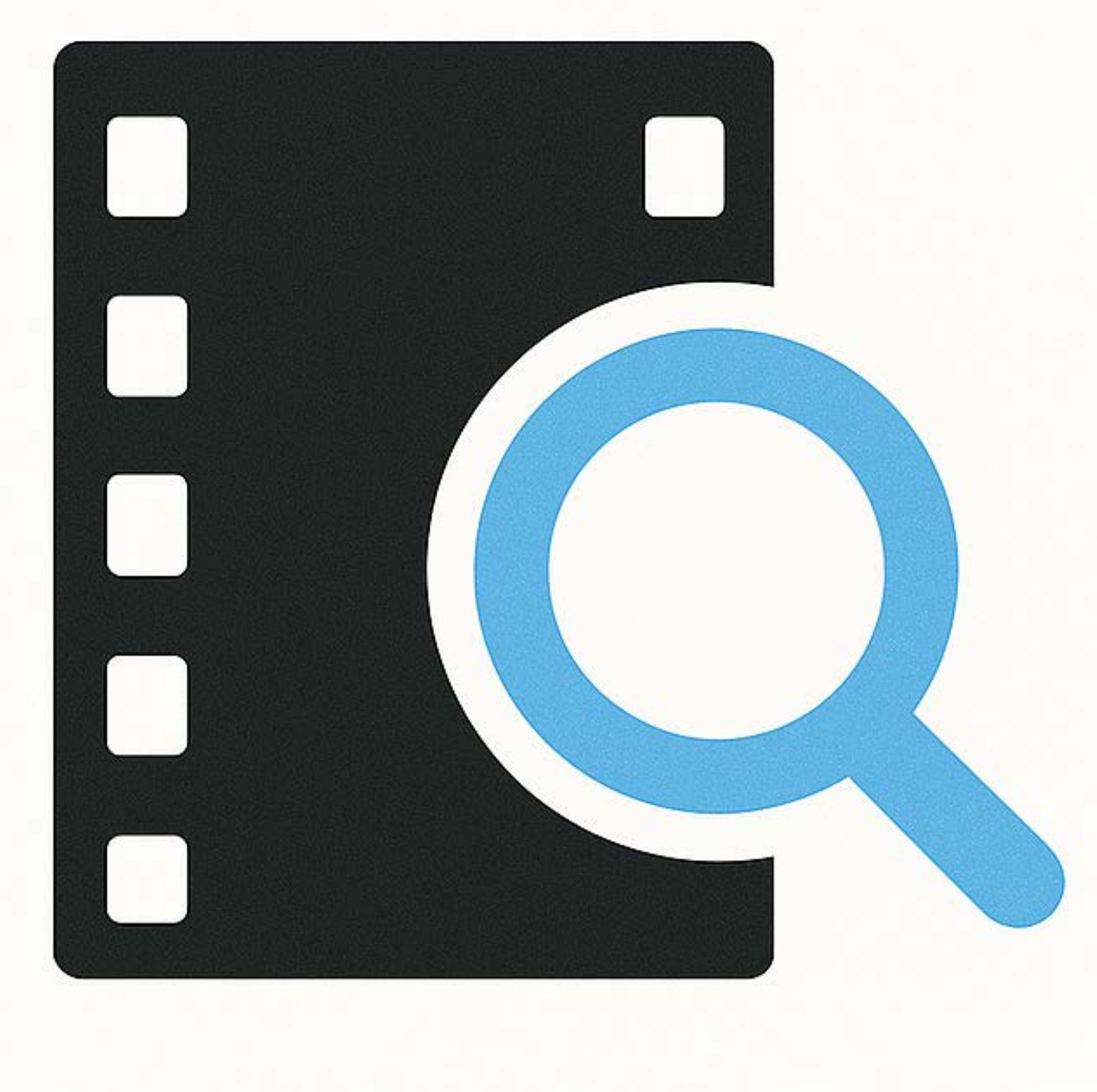} \huge\bfseries \method{}: Large Multimodal Models for Video Understanding and Editing}
\author{Intelligent Editing Team\footnote{A detailed contributor list can be found in Section \ref{sec:contributor}.} , Intelligent Creation, ByteDance Inc. \\ San Jose/Seattle, US \\ \url{https://bytedance.github.io/vidi-website/}}
\date{}

\begin{document}

\maketitle

\input{sections/abstract}

\newpage
\input{sections/introduction}
\input{sections/overview}
\input{sections/architecture}
\input{sections/multimodal_alignment}
\input{sections/supervised_finetuning}
\input{sections/benchmark}
\input{sections/results}
\input{sections/related_work}
\input{sections/conclusion}
\input{sections/contributors}

{
\small
\bibliographystyle{plain}
\bibliography{main}
}
\end{document}

%% file: sections/abstract.tex
\vspace{-1.2cm}
\section*{Abstract}
\small{Humans naturally share information with those they are connected to, and video has become one of the dominant mediums for communication and expression on the Internet. To support the creation of high-quality large-scale video content, a modern pipeline requires a comprehensive understanding of both the raw input materials (\eg, the unedited footage captured by cameras) and the editing components (\eg, visual effects). In video editing scenarios, models must process multiple modalities (\eg, vision, audio, text) with strong background knowledge and handle flexible input lengths (\eg, hour-long raw videos), which poses significant challenges for traditional models. 
In this report, we introduce \method{}, a family of Large Multimodal Models (LMMs) for a wide range of video understand editing scenarios. The first release focuses on temporal retrieval, \ie, identifying the time ranges within the input videos corresponding to a given text query, which plays a critical role in intelligent editing. The model is capable of processing hour-long videos with strong temporal understanding capability, \eg, retrieve time ranges for certain queries. 
To support a comprehensive evaluation in real-world scenarios, we also present the VUE-TR benchmark, which introduces five key advancements: 1) \textbf{Video duration}: spans from $20$ seconds to over an hour, which is significantly longer than existing temporal/moment retrieval datasets. 2) \textbf{Audio support}: includes audio-based queries for temporal retrieval. 3) \textbf{Query format}: accommodates three different query lengths/formats, \ie, keyword, phrase, and sentence. 4) \textbf{Annotation quality}: all ground-truth time ranges are manually annotated with high accuracy. 5) \textbf{Evaluation metric}: a refined IoU metric designed to support evaluation over multiple time ranges. Remarkably, \method{} significantly outperforms leading proprietary models, \eg, GPT-4o and Gemini, on the temporal retrieval task, indicating its superiority in video editing scenarios.}
\begin{figure*}[!hbtp]
    \centering
    \vspace{-0.3cm}
    \includegraphics[trim={35 10 40 0},clip, width=1.\linewidth]{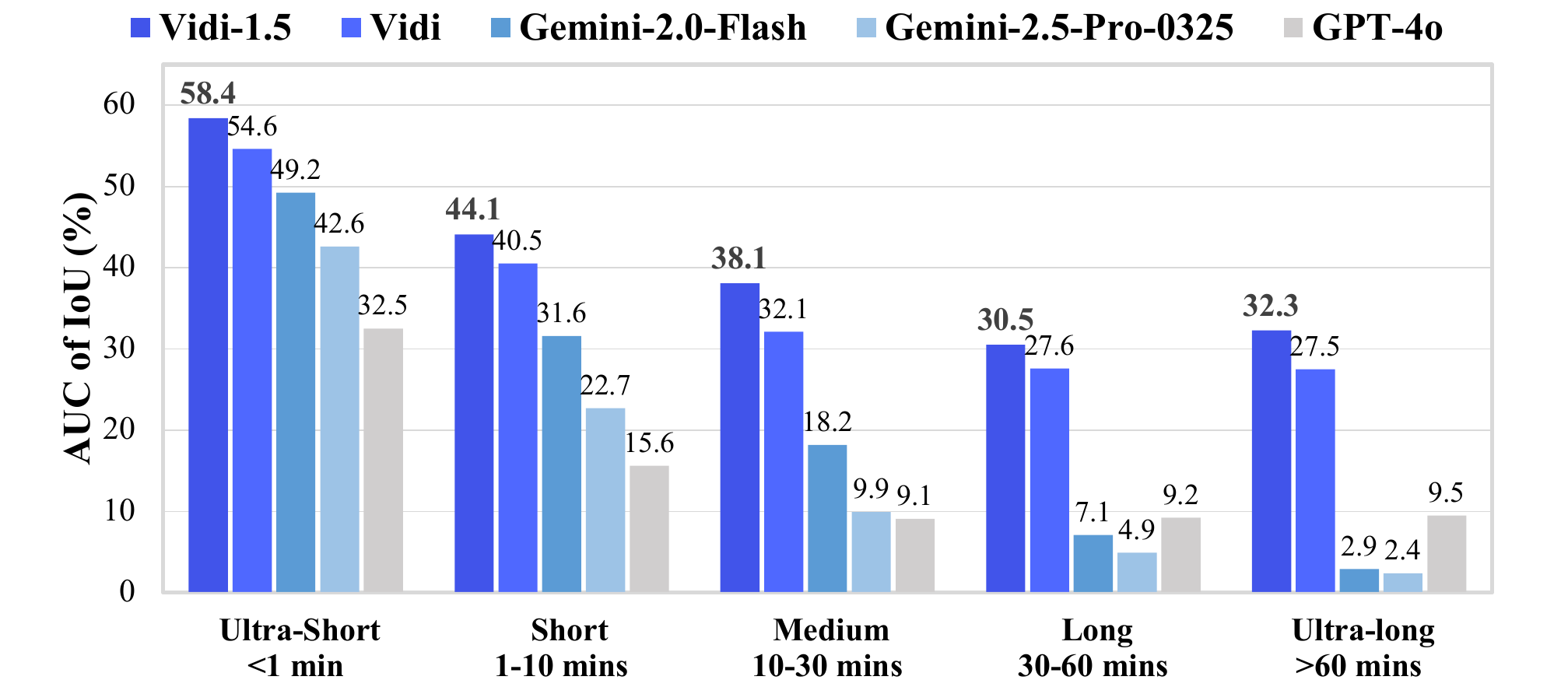}
    \vspace{-0.7cm}
    \caption{Temporal retrieval accuracy of different models on the proposed VUE-TR benchmark.}
    \vspace{-0.4cm}
    \label{fig:teaser}
\end{figure*}

%% file: sections/introduction.tex
\section{Introduction}
Video has become one of the most dominant mediums for sharing information on the Internet. However, for most users, video creation remains a complex and time-consuming process, especially on mobile devices where precise editing is challenging. 
Among all editing stages, the most laborious step is often identifying the desired segments with long, unedited footage.
Beyond trimming, users frequently struggle with video composition tasks, such as selecting appropriate music, transitions, effects, animations, filters, stickers, and fonts, which require both technical skill and artistic judgment.
Moreover, creative editing actions beyond composition are increasingly important in modern video production but are even harder to achieve without expert tools or AI assistance, \eg, generating thumbnails, cover art or stylized scenes. 
\textit{We aim to build the next generation of video creation systems powered by advanced video understanding and editing capabilities, enabling users to complete complex editing workflows automatically and effortlessly.}  

\begin{figure}[!bp]
    \centering
    \vspace{-0.5cm}
    \includegraphics[width=1.\linewidth]{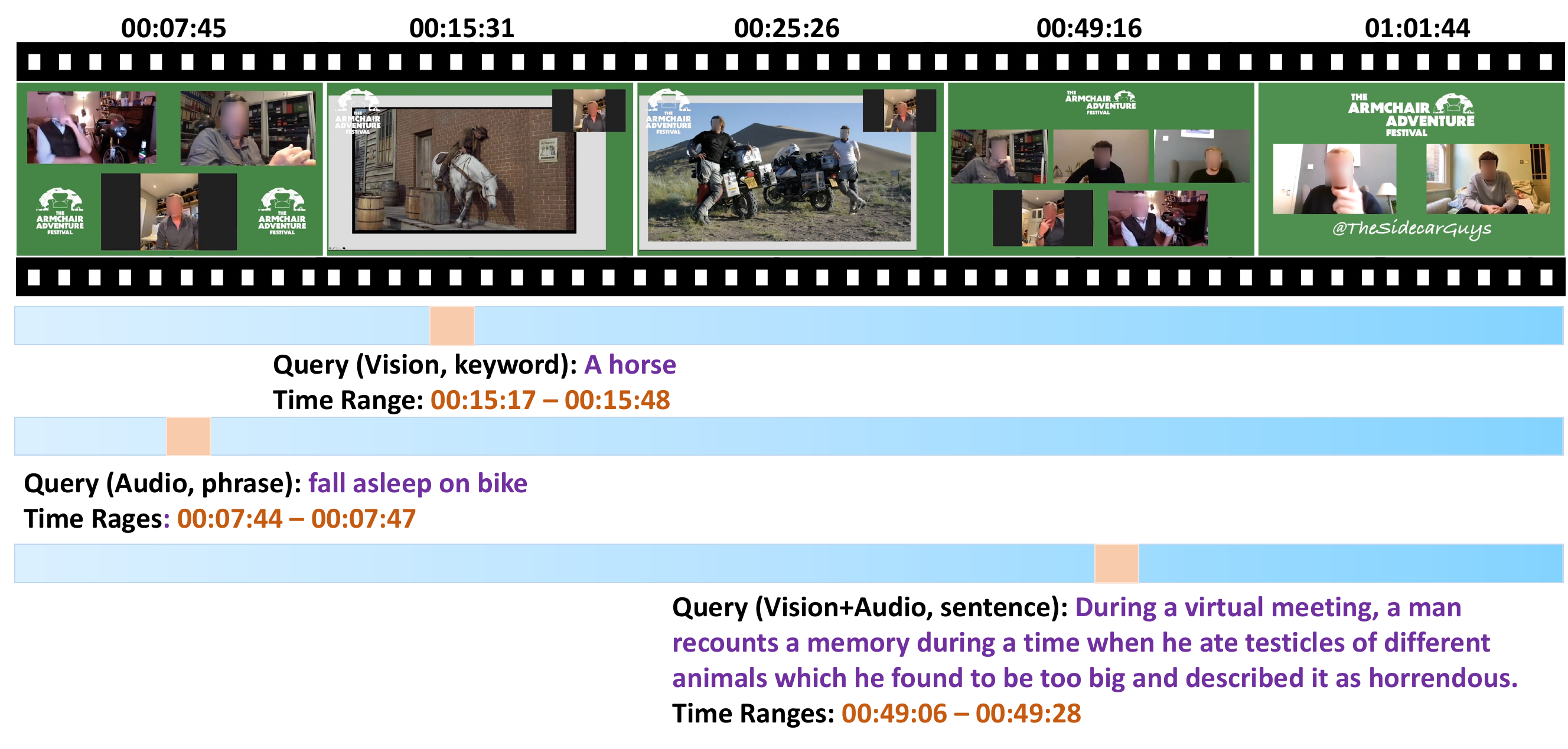}
    \vspace{-0.8cm}
    \caption{Examples of temporal retrieval queries and their corresponding time ranges from the proposed VUE-TR benchmark. The duration of the example video is $3,871$ seconds (\ie, 01:04:31). Each query is presented in one of three formats (\ie, keyword, phrase, sentence) with visual or audio information. Facial regions in the video frames are blurred to protect privacy.}
    \label{fig:intro_fig}
\end{figure}

In this release, we present \method{}, a large multimodal model for video understanding and editing (\textbf{VUE}) with a particular focus on the temporal retrieval (\textbf{TR}) task. 
\method{} takes text, vision, and audio as input modalities and retrieves the most relevant time ranges corresponding to a natural language query. 
A key strength of our model lies in its ability to handle hour-long input videos, significantly surpassing the duration constraints of existing academic temporal/moment retrieval datasets \cite{charades, DBLP:conf/iccv/KrishnaHRFN17, DBLP:conf/iccv/HendricksWSSDR17, qvhighlight, DBLP:journals/corr/abs-2411-19772}, which typically cap at around $150$ seconds. 

To support comprehensive evaluation in realistic settings, we introduce a manually annotated benchmark (abbreviated VUE-TR) for temporal retrieval.
VUE-TR consists of videos ranging from approximately $20$ seconds to over $1$ hour, categorized into five groups: ultra-short ($<1$ min), short ($1-10$ mins), medium ($10-30$ mins), long ($30-60$ mins) and ultra-long ($>1$ hour).
Each video is paired with queries of varying formats and lengths (\ie, keyword, phrase, and sentence) to reflect the diversity of the user search intent.
Critically, queries may require visual, audio, or both modalities for accurate localization. This aligns closely with real-world scenarios, where audio plays an essential role in video comprehension, particularly in domains such as TV shows, broadcasts, and musical performances. 
As illustrated in Figure \ref{fig:intro_fig}, we present an example of an hour-long video with three queries, covering different formats and modalities.
Remarkably, \method{}-7B significantly surpasses leading Large Multimodal Models (LMMs), such as Gemini and GPT-4o, highlighting its effectiveness in handling complex multimodal temporal retrieval tasks. Building upon this success, \method{}-1.5-9B achieves further performance gains through an enhanced LLM backbone, an upgraded visual encoder, an adaptive token compression strategy, and the incorporation of additional high-quality training data.

%% file: sections/overview.tex
\section{General Overview}
\begin{figure}[!htbp]
    \centering
    \includegraphics[width=.7\linewidth]{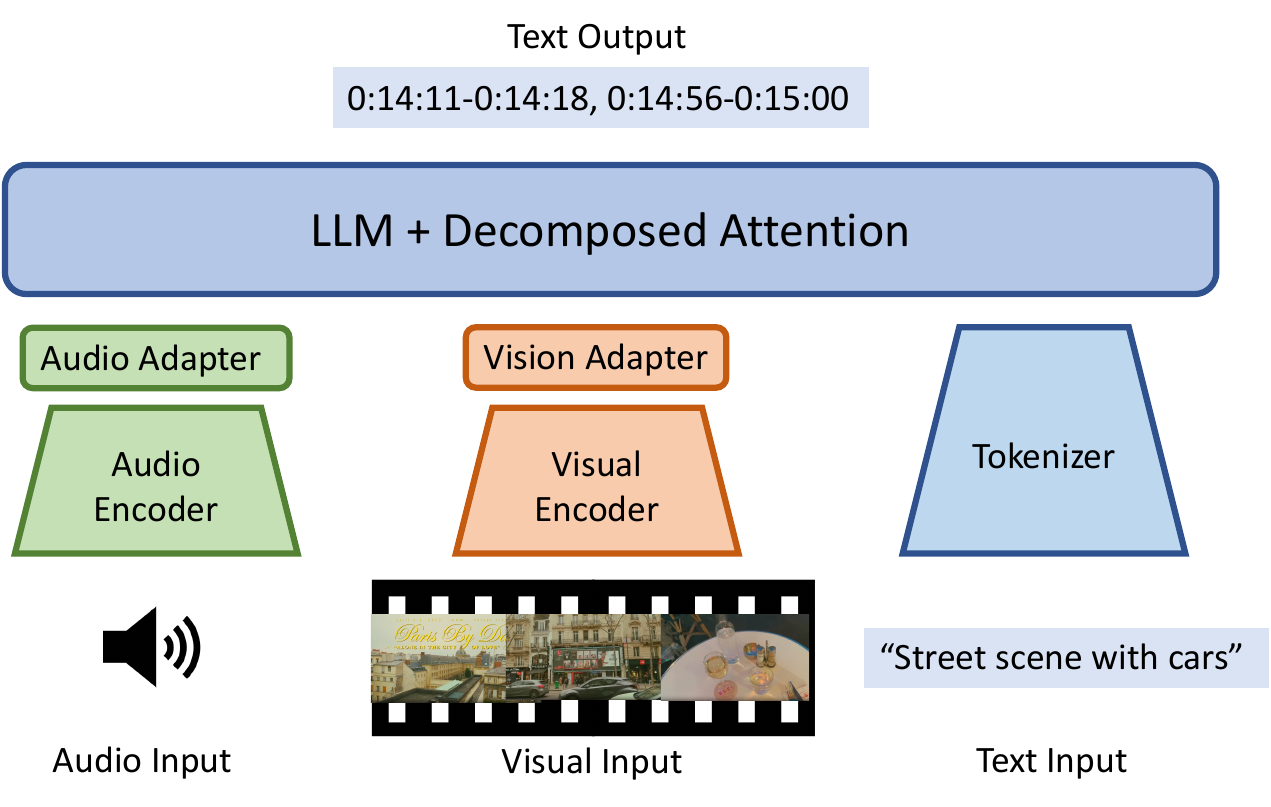}
    \caption{An overview of the \method{} architecture. Raw visual and audio inputs are first process by pretrained modality-specific encoders to extract token sequences. A multimodal adapter is applied to both the visual and audio branches to compress the token sequences and project them into the input space of the pretrained LLM \cite{jiang2023mistral,team2024gemma}. Notably, the LLM operates using the decomposed attention~\cite{kuo2025rethinking} to enable efficient and scalable modeling over long, densely sampled multimodal sequences.}
    \label{fig:overview}
\end{figure}

Figure \ref{fig:overview} illustrates the model architecture of \method{}. For LLM training, we adopt the decomposed attention mechanism proposed by Kuo \etal~\cite{kuo2025rethinking}, which reduces the computational complexity over multimodal tokens $N$ from $\mathcal{O}(N^{2})$ to $\mathcal{O}(N)$ without sacrificing performance on downstream multimodal tasks. This efficient attention design enables training and inference in extremely long videos, which are otherwise infeasible for standard transformer-based models \cite{DBLP:conf/nips/VaswaniSPUJGKP17}.
All videos are uniformly sampled at $1$ frames per second (fps) for visual input and $16,000$ Hz for audio, ensuring that the model can localize and understand content with second-level precision.

Training of \method{} is carried out in two main stages to ensure strong multimodal temporal retrieval:
\begin{itemize}
    \item \textbf{Multimodal Alignment}: This stage focuses on aligning vision and audio data with the corresponding text descriptions and timestamps. We first train the vision and audio adapters using visual and audio captioning data, while keeping the rest of the model frozen. Then the LLM and adapters are jointly trained on synthetic video/audio data with ground-truth time ranges, allowing the model to learn multimodal-to-temporal grounding. Finally, the model is fine-tuned using real videos, with supervision from paired (\ie, timestamp and caption/ASR) data to narrow the domain gap between synthetic and in-the-wild content.
    
    \item \textbf{Application Post-training}: To support diverse application scenarios, we further fine-tune the model on task-specific data, including temporal retrieval and video/image VQA. The temporal retrieval training data mirror the structure of the VUE-TR benchmark, with queries in three formats (\ie, keyword, phrase, sentence), requiring information from vision, audio, or both. This stage enhances the model's ability to handle real-word usage patterns where query types vary significantly.
\end{itemize}
Depending on the intended use case, we train different versions of \method{}, with custom training recipes, \eg, a temporal retrieval expert model, and a more generic VQA model (to be included in future releases). During inference, \method{} is capable of running on a single 80G GPU without quantization, and efficiently process videos exceeding $2$ hours in length, making it practical for real-world deployment in long-form video editing and retrieval tasks. 

\clearpage

%% file: sections/architecture.tex
\section{Model Architecture}
Unlike generic video understanding tasks that rely on sparse or uniform sampling of a fixed number of frames, temporal retrieval requires models to localize relevant segments with second-level precision, necessitating dense sampling, typically at $1$ fps.
Under this fixed sampling rate, the multimodal inputs (\eg, vision, audio, and text) can vary significantly in length, ranging from ultra-short of just a few seconds to ultra-long exceeding an hour.
Furthermore, our proposed VUE-TR benchmark emphasizes realistic retrieval conditions by requiring models to process both visual and audio modalities when localizing queries.
To address challenges in handling videos with varying length and multimodal inputs at scale, we adopt the Decomposed Attention (\textbf{D-Attn})~\cite{kuo2025rethinking} architecture in \method{}.
D-Attn offers exceptional computational efficiency while maintaining strong multimodal capabilities.
It also facilitates seamless integration of visual, audio, and textual streams, making it well suited for temporal retrieval in long-form, real-world videos.

As shown in Figure~\ref{fig:overview}, the visual and audio inputs are first encoded into token sequences using pretrained visual and audio encoders, respectively.
The resulting multimodal embeddings are projected with corresponding adapter layers. Together with the text query embeddings, they are fed into a D-Attn LLM, which localizes the text query within the input video.

\begin{wrapfigure}{r}{0.5\linewidth}
  \begin{center}
    \includegraphics[width=0.8\linewidth]{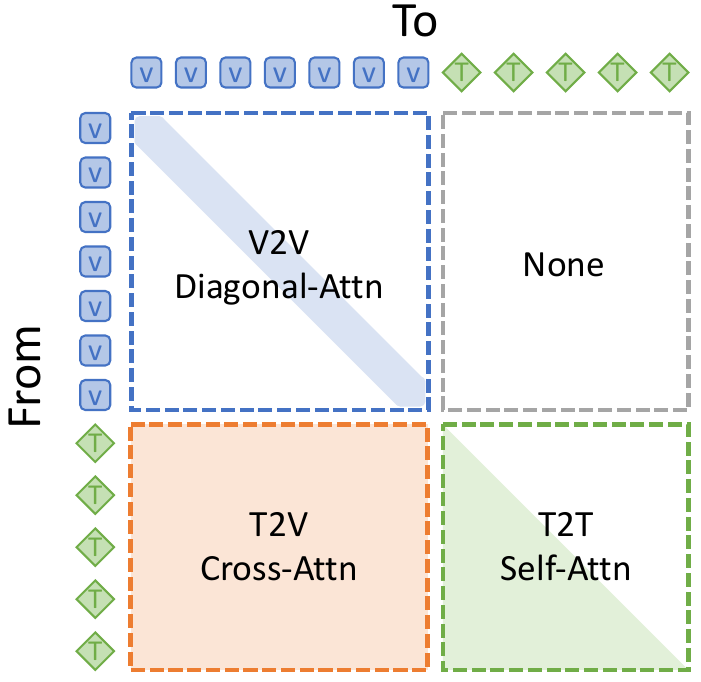}
  \end{center}
  \caption{Decomposed Attention~\cite{kuo2025rethinking} equivalently decomposes causal self-attention in an LLM into three components: visual-to-visual (V2V), textual-to-textual (T2T), and textual-to-visual (T2V) attentions.}\label{fig:decomp-attn}
\end{wrapfigure}

As illustrated in Figure~\ref{fig:decomp-attn}, D-Attn is an architectural adaptation of modern pretrained LLMs, where causal self-attention within an LLM is equivalently decomposed into visual-to-visual self-attention (V2V Self-Attn), textual-to-textual self-attention (T2T Self-Attn), and textual-to-visual cross-attention (T2V Cross-Attn).
With this decomposition strategy, Kuo~\etal~\cite{kuo2025rethinking} introduce novel modifications to both V2V Self-Attn and T2V Cross-Attn to enhance multimodal performance and computational efficiency, while preserving the capabilities of the pretrained LLM.

From a computational point of view, a diagonal variant of V2V Self-Attn is proposed to retain performance while reducing the complexity from $\mathcal{O}(N^2)$ to $\mathcal{O}(N)$ for the visual tokens $N$, shown in Figure~\ref{fig:decomp-attn}. This lightweight design is particularly well suited for densely sampled, ultra-long videos, where $N$ might be extremely large. For example, if each frame consists of 400 tokens, a one-hour video would have 1.44M tokens. Such input length is challenging for conventional fully self-attention mechanisms.

In terms of multimodal alignment, Kuo~\etal~\cite{kuo2025rethinking} identify a critical issue: positional bias between text and visual tokens can hinder the model's ability to establish comprehensive understanding of multimodal contents.
To address this, a debiased positional encoding strategy is proposed to remove positional encodings from T2V Cross-Attn, eliminating the undesirable bias. This approach yields consistent improvements across a wide range of downstream tasks.
By integrating both diagonal V2V Self-Attn and debiased positional encodings in T2V Cross-Attn, the resulting D-Attn LLM can process multimodal tokens longer than the pretrained LLM's original context length. Please refer to Kuo~\etal~\cite{kuo2025rethinking} for more details.

To effectively address the challenge of varying-length video input, we modify the $\alpha$-weighting strategy used in the D-Attn framework. This adjustment is designed to balance the contributions of textual and multimodal information. In particular, video lengths (and thus token counts) can vary significantly in the context of temporal retrieval.
In D-Attn, Kuo~\etal~\cite{kuo2025rethinking} analytically derive how self-attention can be equivalently decomposed into a weighted sum of T2T Self-Attn and T2V Cross-Attn.
Given a text token $t$ and a sequence of concatenated visual and textual tokens $[V, T]$, the attention from $t$ to $[V, T]$ can be formulated as
\begin{equation}
\texttt{Attn}(t, [V, T]) = \alpha_V\ \texttt{XA}(t, V) + \alpha_T\ \texttt{SA}(t,T),
\end{equation}
where $\alpha_V=\texttt{Sigmoid}(S_V - S_T)$, $\alpha_T=\texttt{Sigmoid}(S_T - S_V) = 1-\alpha_V$. $V = [v_1, v_2, \cdots, v_N]$ and $T = [t_1, t_2, \cdots, t_M]$ represent a sequence of visual and text tokens, respectively.
$S_V$ and $S_T$ are defined as the logarithmic sum of the exponential of the dot product between $t$ and $V$, and between $t$ and $T$, respectively.
\begin{equation}
S_V = \log\left(\sum_{n}^{N} e^{\bm{q}_{t}\cdot \bm{k}_{v_n}}\right) \text{, and}\ S_T = \log\left(\sum_{m}^{M} e^{\bm{q}_{t}\cdot \bm{k}_{t_m}}\right),
\end{equation}
where $N$ is the number of visual tokens and $M$ is the number of text tokens. In the temporal retrieval task, this formulation leads to an imbalance: while the length of the text query $M$ typically stays within a small range, the count of visual tokens $N$ can vary dramatically due to differences in the duration of the video. As a result, $S_V$ tends to be significantly larger for longer videos, which in turn biases $\alpha_V$ towards $1$ and $\alpha_T$ towards $0$. This causes the model to overemphasize visual input while neglecting the textual query, potentially degrading performance.
To mitigate this issue, we simplify the formulation in \method{} by fixing the weighting coefficients, \ie, $\alpha_V = \alpha_T = 1$. This ensures a balanced contribution from both modalities, regardless of the input video length.

The D-Attn framework can be trivially generalized to accommodate audio inputs.
Given audio tokens $A=[a_1,a_2,\cdots,a_P]$, the attention of a text token $t$ to the combined multimodal sequence $[V,A,T]$ can be similarly derived as
\begin{align}
\texttt{Attn}(t, [V, A, T]) &= \alpha_V\ \texttt{XA}(t, V) + \alpha_A\ \texttt{XA}(t, A) + \alpha_T\ \texttt{SA}(t,T) \\
& \simeq \texttt{XA}(t, V) + \texttt{XA}(t, A) + \texttt{SA}(t,T),
\end{align}
where we set $\alpha$-weightings $\alpha_V = \alpha_A = \alpha_T = 1$.
In practice, we observe that this fixed-weight decomposition significantly improves training stability, accelerates convergence, and yields better performance than both the original $\alpha$-adaptive formulation and fully self-attention.

%% file: sections/multimodal_alignment.tex
\section{Multimodal Alignment}
To align the multimodal inputs (vision and audio) with the corresponding timestamps and text in either the inputs or outputs, we adopt a three-stage training strategy: 1) adapter training, 2) synthetic data training, and 3) real video training. 

\subsection{Adapter Training}
In this stage, we train the adapters from scratch while keeping the weight of the vision/audio encoders and the LLM fixed. Each adapter contains a convolution layer for compressing the raw visual or audio tokens, followed by an MLP that maps the compressed representations into the LLM-compatible input space. We leverage the strong publicly available pretrained models for all other components: SigLIP series \cite{zhai2023sigmoid, DBLP:journals/corr/abs-2502-14786} for vision, Whisper \cite{radford2023robust} for audio, and Mistral \cite{jiang2023mistral}/Gemma \cite{team2024gemma} for the language model backbone. The adapters are trained on approximately $1$ million image and audio caption data, allowing them to learn semantic grounding across modalities while maintaining alignment with the LLM. 

\subsection{Synthetic Data Training}
\begin{figure*}[!htbp]
    \centering
    \includegraphics[width=\linewidth]{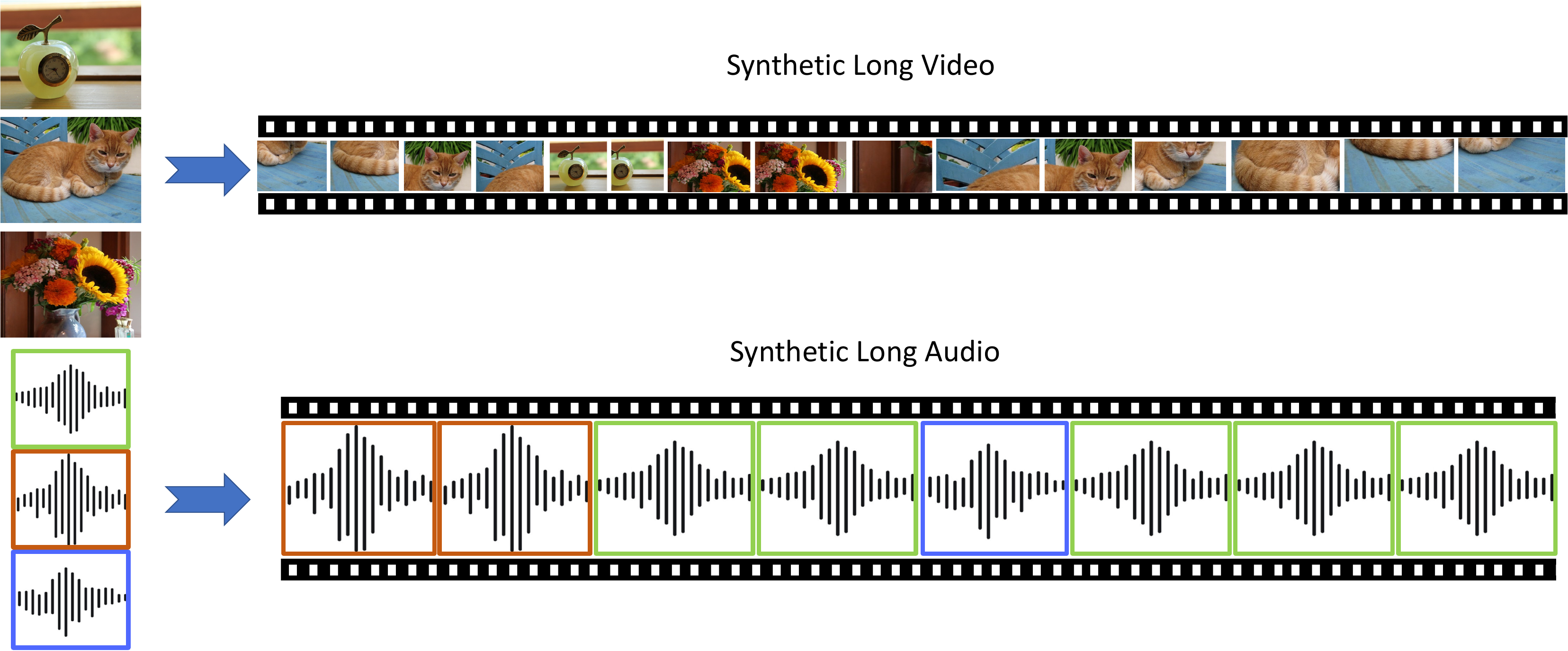}
    \caption{An overview of the synthetic training data generation pipeline. Visual and audio segments can be duplicated at multiple timestamps to simulate real videos.}
    \label{fig:synthetic}
\end{figure*}
To support large-scale training for temporal alignment, we propose a synthetic video generation pipeline based on captioned video and audio datasets with ground-truth timestamps.  As shown in Figure \ref{fig:synthetic}, we begin by randomly sampling a group of candidate samples from large-scale caption corpora, \ie, approximately $25$ million images and $400$ thousand audio clips.
Each image is expanded into a synthetic video segment using a sliding window approach with randomized parameters such as window size, starting corner, sliding direction, and sliding speed. This mimics camera movement or visual variation over time.
Similarly, audio clips are segmented and spliced in random order to create synthetic audio tracks. The resulting synthetic inputs are arbitrarily long and composed of diverse multimodal segments.
Since each segment originates from a known captioned sample, we can automatically generate large-scale (timestamp, caption) pairs as supervision.
Since images do not match audio, the training input contains only one active modality (vision or audio), while the other is padded with empty tokens.

We train the model on two complementary objectives: 1) caption prediction: given a timestamp range, predict the corresponding caption; 2) timestamp localization: given a caption, predict the matching timestamp range.
These dual tasks help the model develop a deep understanding of both visual/audio content and the temporal axis. After training on synthetic data, the model achieves over $80\%$ accuracy on a synthetic evaluation set with multi-minute sequences, indicating strong alignment capability and robust temporal grounding.

\subsection{Real Video Training}
\label{sec:real}
\begin{figure*}[!htbp]
    \centering
    \includegraphics[width=\linewidth]{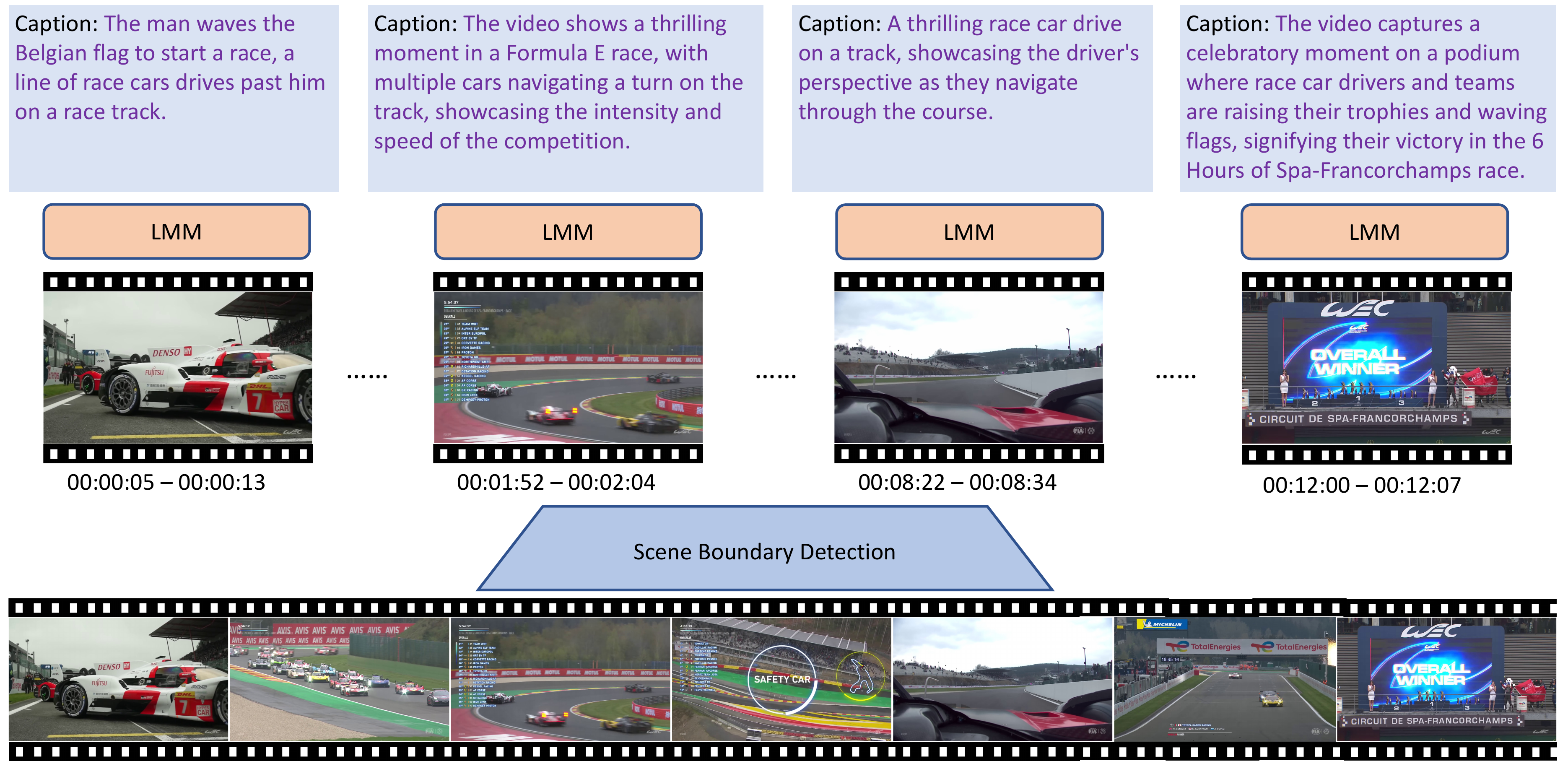}
    \caption{An overview of the proposed real video training data generation pipeline. Long videos are first segmented into short clips using scene boundary detection and subtitle punctuation. Then, existing LMMs are used to generate dense, timestamp-aligned captions for each clip, resulting in high-quality supervision for real-world temporal grounding tasks.}
    \label{fig:realvideo}
\end{figure*}

To bridge the gap between synthetic videos and real-world videos, we further train the model on a large corpus of $1$ million long videos, each annotated with dense supervision in the form of pairs (timestamp, caption) and (timestamp, subtitle).
As shown in Figure \ref{fig:realvideo}, each long video is first segmented into short clips (typically $5-30s$) using a combination of scene boundary detection \cite{li2022structured} and subtitle punctuation cues \cite{guhr-EtAl:2021:fullstop}. This process yields approximately $3-500$ segments per video, depending on the density and duration of the content. 
After that, dense captions are generated with state-of-the-art LMMs \cite{bai2025qwen2, zhang2023internlm}, resulting in fine-grained semantic coverage. Original subtitles are reprocessed into sentence-level units using punctuation-based splitting, enhancing their suitability for timestamped supervision. 
This pipeline produces more than $30$ million paired training samples (timestamp, caption) and (timestamp, subtitle). The training input consists of raw video files containing both visual and audio tracks.

We design four training objectives to fully leverage aligned supervision.
\begin{itemize}
    \item Caption prediction: given a timestamp range, predict the corresponding dense caption.
    \item Subtitle prediction: given a timestamp range, predict the associated sentence-level subtitle.
    \item Caption-based localization: given a caption, predict the matching timestamp range.
    \item Subtitle-based localization: given a sentence-level subtitle, predict the corresponding timestamp range.
\end{itemize}
These tasks not only reinforce the model's ability to align textual descriptions with multimodal signals, but also help it adapt to complex video/audio in real-world distributions.

%% file: sections/supervised_finetuning.tex
\section{Application Post-training}
\label{sec:annotation}
\begin{figure}[!htbp]
    \centering
    \includegraphics[width=\linewidth]{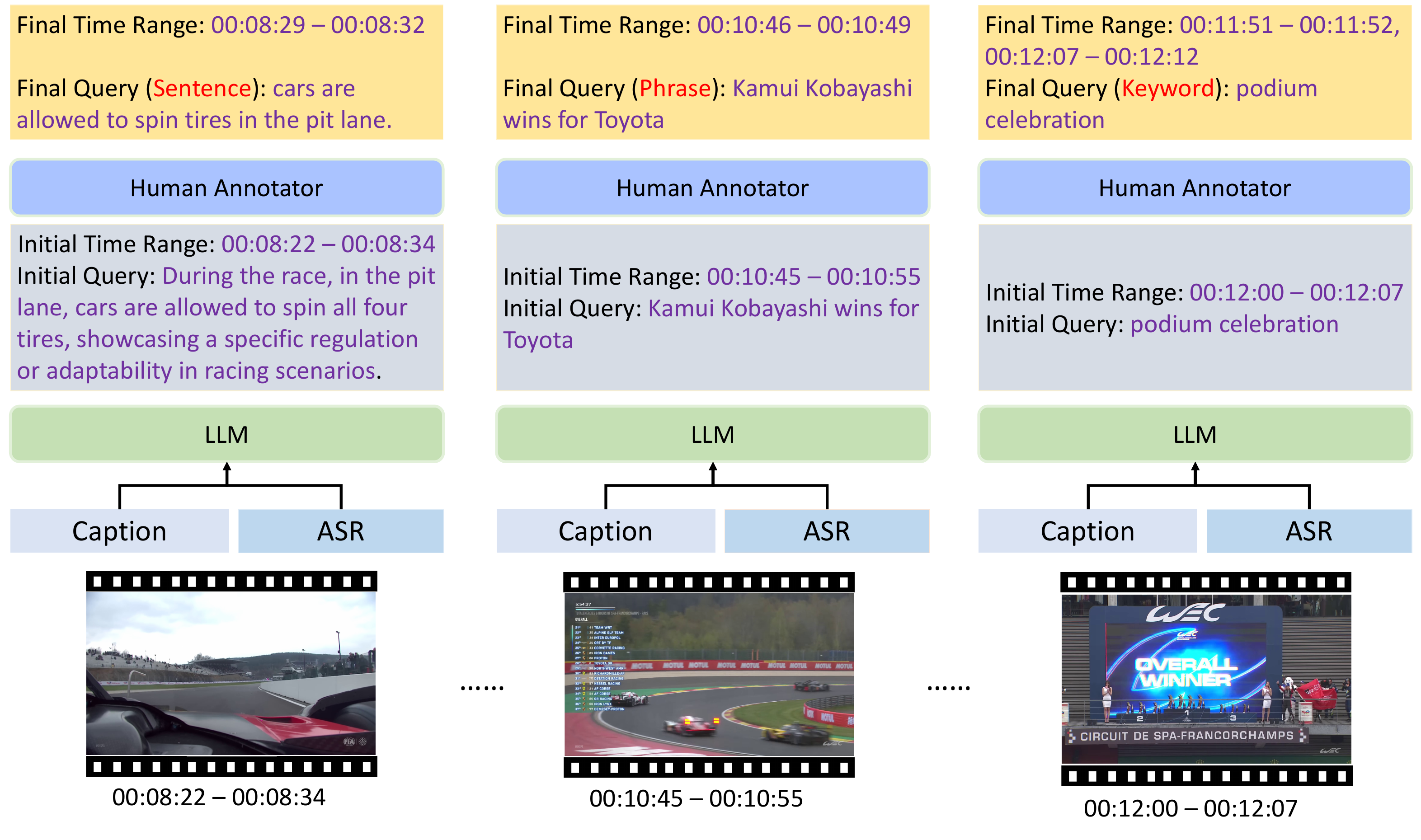}
    \caption{An illustration of post-training data generation pipeline for temporal retrieval. Queries of varying formats (keywords, phrases, and sentences) are constructed and paired with timestamp annotations from real videos.}
    \label{fig:annotation}
\end{figure}

To support the temporal retrieval task, we build an annotation pipeline to generate user-like search queries and ground-truth timestamps to guide the training phase. As shown in Figure \ref{fig:annotation}, we take advantage of the video clip split in Section \ref{sec:real} with dense captions to generate user-style queries and timestamp ranges. 

{\flushleft \textbf{Dense Captions.}} Similar to \cite{DBLP:journals/corr/abs-2410-08260}, we generate structural captions to maintain the consistency of text-clip and cover the details in each video clip. Specifically, each caption contains six aspects: 1) subjects of the video, 2) actions of the subjects, 3) scene environment where the action takes place and all visual text overlay on-screen, including logos, subtitles, signs, or other writings, 4) visual style or special effects including video effects, animation, style, composition, and lighting, 5) camera parameters including camera motion, angles, and focal length, 6) background knowledge for common sense reasoning such as famous landmarks, celebrities, or historical events.

{\flushleft \textbf{Query and Timestamp Pairs.}}  To generate high-quality query–timestamp pairs for long videos, we use chain-of-thought (CoT) \cite{DBLP:conf/nips/Wei0SBIXCLZ22} prompting of LLM.
We first combine detailed dense captions and subtitles with timestamps from all video segments to provide a comprehensive textual representation of the video's content. The aggregated text is sent to a pretrained LLM. The LLM is used to generate a concise summary to capture the key events and themes with the guidance from the CoT prompting to comprehend the context of the video. In this way, CoT prompting enables the model to perform intermediate reasoning, leading to a more accurate and coherent understanding of video content. 

To improve format diversity and better simulate real-word scenarios, we generate queries in three levels of granularity:
\begin{itemize}
    \item \textbf{keyword}: concise terms representing objects, concepts or entities, such as ``love'', ``coffee making process'', and ``washing machine''.
    \item \textbf{phrase}: short descriptions capturing actions or states, like ``a man riding a bike'', ``person in deep thought'', and ``enjoying a swim in the pool''.
    \item \textbf{sentence}: complete sentence(s) describing detailed events or scenes, \eg, ``The majestic presence of a volcano surrounded by lush vegetation and shrouded in clouds''.
\end{itemize} 

{\flushleft \textbf{Post-processing and Filtering.}}
We design a rule-based post-processing step to enhance the quality of generated queries and the corresponding time ranges, formed by the following stages.
1. Merging adjacent time ranges: for each query, we combine consecutive time segments or those separated by small gaps (\eg, $0.5$ seconds), provided the query appearing in captions or subtitles.
2. Filtering out low-confidence queries: queries with confidence scores below $0.9$ are discarded to ensure reliability.
3. Excluding overly general queries: queries associated with more than $10$ timestamps are considered too general. We remove such cases that frequently appear throughout the video.
4. Eliminating machine-style queries: queries exhibiting templated or unnatural phrasing, such as ``The video concludes...'' or ``In the closing moments,'' will be filtered out to ensure quality.
After that, the generated queries and the corresponding time ranges are sent to human annotators to perform the final stage verification and modification.

{\flushleft \textbf{Two-Round Human Annotation.}}
Although involving a filtering process, generated queries and the corresponding time ranges still contain ambiguity of phrasing and misaligned timestamps. To obtain high-quality annotations, we further design a comprehensive human-in-the-loop refinement process.
Annotators begin by examining each query for clarity and relevance. If a query is ambiguous or unrealistic, it is required to be rewritten to reflect the video content accurately. After that, the annotators review the entire video to identify precise time segments corresponding to the refined queries. In addition, another annotator is required to review the annotations independently. Discrepancies identified during this phase will be resolved through discussion or by consulting a senior annotator.

It is worth mentioning that each query is annotated based on its reliance on visual and/or auditory information. Annotators assign one of the following modality tags to each query: vision, audio, or vision+audio.
\begin{itemize}
    \item \textbf{Vision}. The query can be accurately identified purely on the basis of visual content. Audio information is not necessary to retrieve the segments.
    \item \textbf{Audio}. The query depends solely on the auditory information. Visual cues are not necessary to identify the video segment.
    \item \textbf{Vision+Audio}. The query relies on visual and auditory information for accurate interpretation and localization within the video.
\end{itemize}
This categorization strategy helps researchers evaluate the capabilities of the models with a more detailed analysis. 

%% file: sections/benchmark.tex
\section{Evaluation Benchmark}
We introduce VUE-TR, a new evaluation benchmark specifically designed to advance temporal retrieval in real-world scenarios. VUE-TR addresses five critical aspects often overlooked in prior academic benchmarks~\cite{charades, DBLP:conf/iccv/KrishnaHRFN17, DBLP:conf/iccv/HendricksWSSDR17, qvhighlight, DBLP:journals/corr/abs-2411-19772}: \textbf{video duration}, \textbf{audio support}, \textbf{query format}, \textbf{annotation quality}, and \textbf{metric design}. 
To ensure high-quality supervision, all annotations are manually curated using a robust annotation pipeline described in Section~\ref{sec:annotation}, which yields significantly more accurate and consistent labels than existing benchmarks.
\begin{table}[!htbp]
    \centering
    \resizebox{\linewidth}{!}{
    \begin{tabular}{c|c c c c c |c }
    \toprule[1.5pt]
    \multirow{2}{*}{Duration Category} & Ultra-short & Short & Medium & Long & Ultra-long & \multirow{2}{*}{Total}\\
     & $<1$ min & $1-10$ mins & $10-30$ mins & $30-60$ mins & $>60$ mins \\
     \hline
    \# Videos & 63 & 150 &  150 & 50 & 15 & 428 \\
    \# Queries & 183 & 439 & 427 & 396 & 153 & 1,598\\
    Video Hours  & 0.81 & 11.71 & 43.71 & 34.17 & 17.48 & 107.87 \\
    \bottomrule[1.5pt]
    \end{tabular}}
    \caption{Duration distribution of videos in the proposed VUE-TR evaluation benchmark. The dataset covers a wide range of video lengths, from ultra-short clips ($<1$ minute) to ultra-long videos ($>1$ hour), enabling comprehensive evaluation of temporal retrieval models across diverse real-world scenarios.}
    \label{tab:stats}
\end{table}

\begin{figure}[!htbp]
    \centering
    \includegraphics[width=0.48\linewidth]{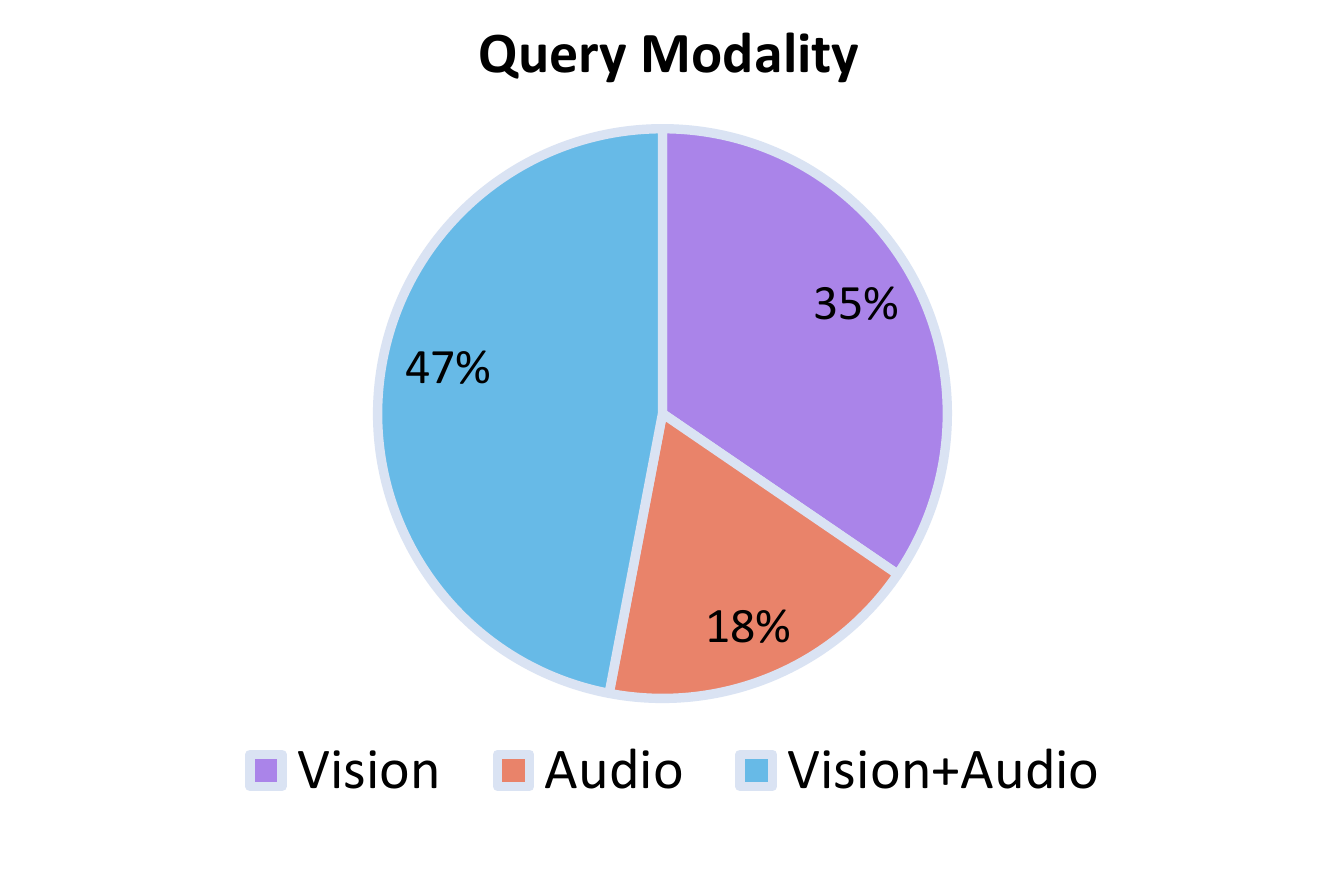}
    \includegraphics[width=0.48\linewidth]{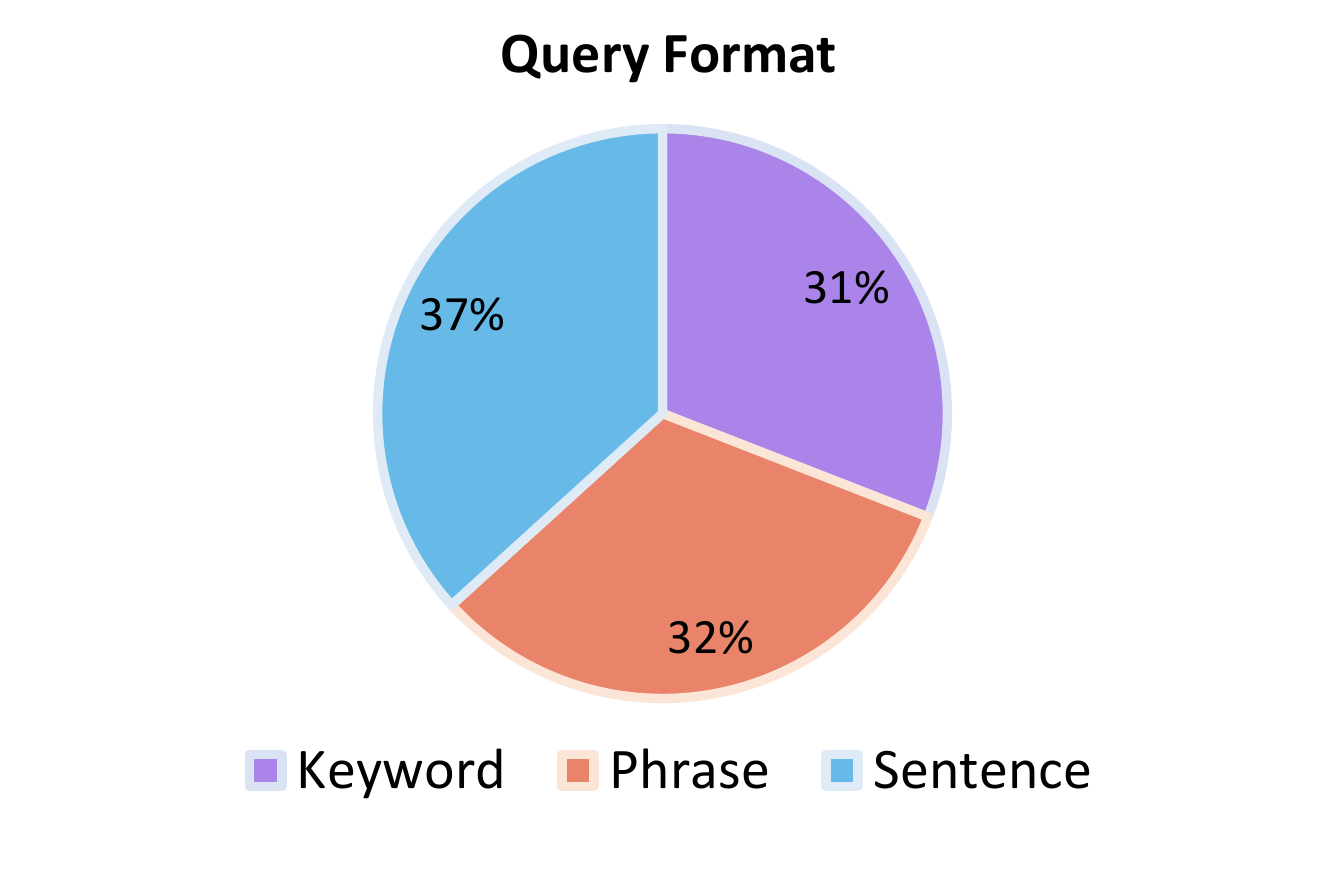}
    \vspace{-0.4cm}
    \caption{The distribution of query modality and format in the VUE-TR benchmark. This diverse distribution reflects real-world retrieval scenarios and enables fine-grained analysis of model performance across input types.}
    \label{fig:distribution}
\end{figure}

\subsection{Data Statistics}
\label{sec:stats}
VUE-TR is built using publicly available videos, and the annotations are made accessible to foster open research. As presented in Table \ref{tab:stats} and Figure \ref{fig:distribution}, the benchmark consists of $1,598$ queries across $428$ videos, spanning over $107$ hours. It supports attribute-based slicing for fine-grained performance analysis as follows.
\begin{itemize}
    \item \textbf{Video Duration.} 
    Unlike prior datasets limited to short clips, the video length varies from $20$ seconds to more than $1$ hour, covering the full spectrum of durations encountered in real-world scenarios. To facilitate duration-wise evaluation, we categorize videos into five balanced buckets: ultra-short ($<1$ min), short ($1-10$ mins), medium ($10-30$ mins), long ($30-60$ mins) and ultra-long ($>60$ mins). It enables systematic analysis of model performance as the video length increases, which is a key challenge for temporal retrieval that prior work failed to capture.

    \item \textbf{Query Modality.} 
    While most large vision-language models (LVLMs) operate on vision and text alone, VUE-TR explicitly integrates audio as a core input modality.
    As shown in Figure \ref{fig:distribution} (left), $47\%$ of the queries involve both visual and audio signals, $35\%$ are vision-only, and $18\%$ are audio-only. This breakdown enables a comprehensive evaluation of multimodal capabilities.
    
    \item \textbf{Query Format.} 
    VUE-TR is the first temporal retrieval benchmark to incorporate multiple query formats, reflecting the diversity of user intent in real-word search scenarios. The three formats, ranging from keywords to free form sentences, are carefully balanced, as shown in Figure \ref{fig:distribution} (right). Then the model can be evaluated to handle queries of varying linguistic complexity.
\end{itemize}

\subsection{Evaluation Metric}
\label{sec:eval}
\begin{figure*}[!htbp]
    \centering
    \includegraphics[width=\linewidth]{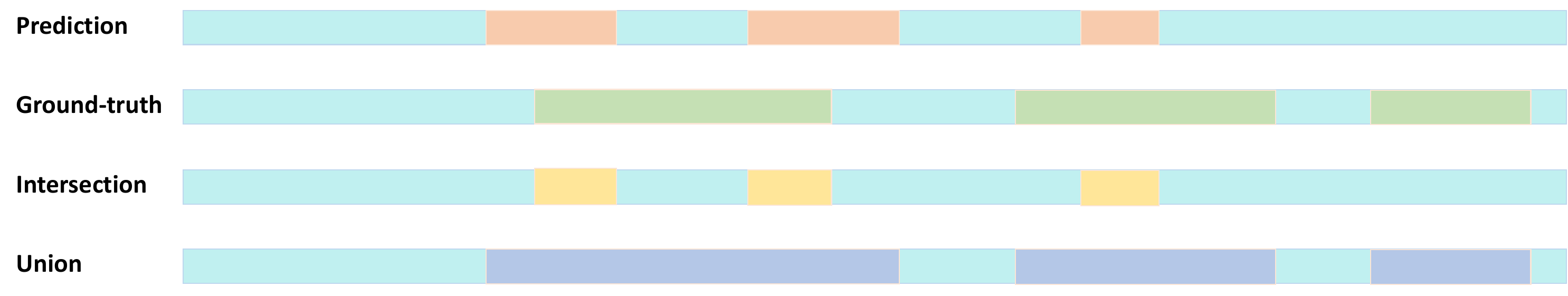}
    \caption{Definition of intersection and union for temporal retrieval. Both prediction and ground-truth annotations may contain multiple timestamp ranges. The IoU for each sample is computed as $\sum \mathcal{I}(T_P,T_G) / \sum \mathcal{U}(T_P,T_G)$.}
    \label{fig:evaluation}
\end{figure*}

To support evaluation of generic temporal retrieval involving multiple timestamp ranges, we re-define \textbf{IoU} (Intersection over Union), \textbf{precision}, and \textbf{recall} along the time axis. As illustrated in Figure \ref{fig:evaluation}, the IoU is computed based on the intersection and the union between the predicted time intervals and ground-truth.

For each sample, we compute precision $P$, recall $R$ and IoU as the core evaluation metrics. According to the definition in Equation \eqref{eq:metric}, a perfect prediction would be exactly the same as the ground-truth timestamp ranges, \ie, $P=R=\text{IoU}=1$. 
To capture performance across varying levels of overlap, we sweep thresholds in the range $[0, 1]$ to generate accuracy-threshold curves. We then compute the area under curve (AUC) for each metric, denoted as $\bar{P}$, $\bar{R}$, $\bar{\text{IoU}}$, which serve as the final evaluation scores. 
Although AUC provides a comprehensive summary, the user experience in real-world applications often depends on performance as specific thresholds, \eg, IoU@0.5. Therefore, we also report accuracy-threshold curves (see Figure \ref{fig:curves}) to facilitate a more detailed analysis, while treating AUC as the primary metric for evaluation.
\begin{equation}
    \left\{\begin{aligned}
    P &= \sum \mathcal{I}(T_P,T_G)  / \sum T_P,  \\
    R &= \sum \mathcal{I}(T_P,T_G)  / \sum T_G,  \\
    \text{IoU} &= \sum \mathcal{I}(T_P,T_G) / \sum \mathcal{U}(T_P,T_G), 
    \end{aligned}\right.
    \label{eq:metric}
\end{equation}
where $\mathcal{I}(\cdot,\cdot)$ and $\mathcal{U}(\cdot,\cdot)$ denote the interaction and union function between predicted timestamps $T_P$ and ground-truth $T_G$, respectively. The summation is taken over all overlapping time intervals between prediction and ground-truth.

%% file: sections/results.tex
\section{Experiment}
\subsection{Implementation Details}
\input{sections/tables/hyperparameters}

We implement the released \method{} models using three core components: the visual encoder (\texttt{google/siglip\\-so400m-patch14-384} \cite{zhai2023sigmoid} for \method{}, and \texttt{google/siglip2-so400m-patch14-384} \cite{DBLP:journals/corr/abs-2502-14786} for \method{}-1.5, the audio encoder \texttt{openai/whisper-large-v3} \cite{radford2023robust}, and the language model ( \texttt{mistralai/Mistral-7B-Instruct-v0.3} \cite{jiang2023mistral} for \method{}, and \texttt{google/gemma-2-9b} \cite{team2024gemma} for \method{}-1.5).
Unlike existing video LLMs that use a fixed number of uniformly sampled frames, \method{} adopts modality-specific fixed sampling rates: $1.0$ fps for video frames and $16,000$ Hz for audio signals. It ensures sufficient coverage of fine-grained visual and auditory details across varying video durations.
After sampling, each image frame is independently encoded by SigLip to convert it into a sequence of visual tokens. In parallel, the audio waveform is segmented into batches, each independently encoded by Whisper, and then concatenated back into a sequence of audio tokens. The complete set of training configurations and hyper-parameters for various stages is provided in Table~\ref{tab:hyperparams}.

\begin{table}[!htbp]
 \small
    \centering
    \resizebox{\linewidth}{!}{
    \begin{tabular}{l c | c c c c c }
    \toprule[1.2pt]
       Category & Metric & \quad\textbf{\method{}}-1.5 & \quad\quad\textbf{\method{}}\quad\quad\quad & Gemini-2.0-Flash$\dagger$ & Gemini-2.5-Pro$\dagger$ & GPT-4o$\ddagger$ \\
        \hline
        \multirow{3}{*}{Ultra-Short ($<1m$)} & $\bar{P}$ & 70.7 & 64.5 & \textbf{72.3} & 58.8 & 53.6\\
        ~ & $\bar{R}$ & 76.2 & \textbf{79.6} & 65.4 & 69.2 & 42.1 \\
        ~ & $\bar{\text{IoU}}$ & \textbf{58.4} & 54.6 & 49.2 & 42.6 & 32.5 \\
        \hline
        \multirow{3}{*}{Short ($1-10m$)} & $\bar{P}$ & \textbf{60.5} & 57.4 & 51.7 & 43.3 & 30.0 \\
        ~  & $\bar{R}$ & \textbf{63.4} & 55.8 & 50.6 & 41.7 & 26.5 \\
        ~  & $\bar{\text{IoU}}$ & \textbf{44.1} & 40.5 & 31.6 & 22.7 & 15.6 \\
        \hline
        \multirow{3}{*}{Medium ($10-30m$)} & $\bar{P}$ & \textbf{53.8} & 47.4 & 35.1 & 31.1 & 18.2\\
        ~ &  $\bar{R}$ & \textbf{54.0} & 46.6 & 34.4 & 22.6 & 21.1\\
        ~ &  $\bar{\text{IoU}}$ & \textbf{38.1} & 32.1 & 18.2 & 9.9 & 9.1 \\
        \hline
        \multirow{3}{*}{Long ($30-60m$)} & $\bar{P}$ & \textbf{47.8} & 38.9 & 26.3 & 29.9 & 18.2\\
        ~ & $\bar{R}$ & 40.6 & \textbf{44.9} & 12.9 & 11.9 & 17.7\\
        ~ & $\bar{\text{IoU}}$ & \textbf{30.5} & 27.6 & 7.1 & 4.9 & 9.2\\
        \hline
        \multirow{3}{*}{Ultra-Long ($>60m$)} & $\bar{P}$ & \textbf{46.1} & 36.7 & 17.1 & 25.3 & 20.4 \\
        ~ &  $\bar{R}$ & 43.2 & \textbf{46.7} & 8.3 & 5.8 & 19.6 \\
        ~ &  $\bar{\text{IoU}}$ & \textbf{32.3} & 27.5 & 2.9 & 2.4 & 9.5 \\
        \hline\hline
        \multirow{3}{*}{Keyword} & $\bar{P}$ & \textbf{60.5} & 54.4 & 50.3 & 47.9 & 37.4\\
        ~ &  $\bar{R}$ & \textbf{61.5} & 57.4 & 33.0 & 33.4 & 26.4\\
        ~ &  $\bar{\text{IoU}}$ & \textbf{43.8} & 39.4 & 21.9 & 17.4 & 18.3 \\
        \hline
        \multirow{3}{*}{Phrase}  & $\bar{P}$ & \textbf{53.1} & 45.3 & 41.4 & 41.7 & 25.1\\
        ~ &  $\bar{R}$ & \textbf{50.2} & 48.1 & 36.6 & 31.1 & 23.7\\
        ~ &  $\bar{\text{IoU}}$ & \textbf{36.8} & 32.4 & 22.3 & 16.0 & 13.2\\
        \hline
        \multirow{3}{*}{Sentence}  & $\bar{P}$ & \textbf{52.9} & 47.5 & 31.1 & 30.2 & 17.5 \\
        ~ &  $\bar{R}$ & \textbf{53.1} & \textbf{52.3} & 34.2 & 23.3 & 22.2 \\
        ~ &  $\bar{\text{IoU}}$ & \textbf{38.5} & 34.7 & 19.7 & 12.6 & 10.0 \\
        \hline\hline
        \multirow{3}{*}{Audio}  & $\bar{P}$ & \textbf{38.1} & 34.2 & 32.5 & 30.9 & 7.3\\
        ~ &  $\bar{R}$ & 42.1 & \textbf{43.8} & 35.8 & 25.5 & 16.5\\
        ~ &  $\bar{\text{IoU}}$ & \textbf{27.0} & 24.2 & 20.3 & 10.2 & 3.7 \\
        \hline
        \multirow{3}{*}{Vision}  & $\bar{P}$ & \textbf{70.0} & 53.6 & 48.6 & 48.2 & 41.3\\
        ~ &  $\bar{R}$ & \textbf{66.9} & 61.6 & 38.9 & 37.9 & 30.5\\
        ~ &  $\bar{\text{IoU}}$ & \textbf{49.7} & 43.9 & 25.4 & 21.8 & 22.0\\
        \hline
        \multirow{3}{*}{Vision+Audio}  & $\bar{P}$ & \textbf{56.2} & 51.2 & 37.2 & 33.9 & 21.2 \\
        ~ &  $\bar{R}$ & \textbf{50.6} & 49.0 & 30.9 & 23.4 & 22.0 \\
        ~ &  $\bar{\text{IoU}}$ & \textbf{36.9} & 33.4 & 18.4 & 12.1 & 11.1 \\       
        \hline\hline
        \multirow{3}{*}{Overall} & $\bar{P}$ & \textbf{55.3} & 49.0 & 40.3 & 39.3 & 25.9 \\
        ~ &  $\bar{R}$ & \textbf{54.8} & 52.5 & 34.6 & 28.9 & 24.0 \\
        ~ &  $\bar{\text{IoU}}$ & \textbf{39.6} & 35.4 & 21.2 & 15.2 & 13.6 \\
    \bottomrule[1.2pt]
    \end{tabular}}
    \caption{
    Performance of different models on the VUE-TR benchmark across various evaluation attributes. $\bar{P}$ and $\bar{R}$ represent the AUC (Area Under Curve) values for precision and recall, respectively; while $\bar{\text{IoU}}$ denotes the AUC of intersection-over-union between prediction and ground-truth timestamp ranges, as defined in Section \ref{sec:eval}. GPT models are accessed via the Azure API, and Gemini is accessed via internal Google API. $\dagger$ Gemini models are evaluated by directly uploading videos. To comply with the $100$ MB upload limit, long videos are resized to a resolution of $256$ pixels.
    Compared to Gemini-2.0-Flash, Gemini-2.5-Pro (0325) incurs higher latency and token cost for reasoning and exhibits a higher content filtering rate, often resulting in empty outputs. $\ddagger$ GPT-4o is constrained by the Azure API's $120$-frame limit. For videos longer than $120$ seconds, we uniformly sample $120$ frames.}
    \label{tab:tr_results}
\end{table}

\subsection{Temporal Retrieval Results}
We compare our model with three state-of-the-art proprietary models including GPT-4o \cite{DBLP:journals/corr/abs-2410-21276}, Gemini-2.0-Flash \cite{DBLP:journals/corr/abs-2312-11805} and Gemini-2.5-Pro \cite{DBLP:journals/corr/abs-2312-11805}. There models are chosen for their strong performance and broad modality support, making them competitive baselines for real-world temporal retrieval.

Since GPT-4o API does not support audio, we extract frames at $1$ fps and feed them as input images. To adapt GPT-4o for temporal retrieval, we design a simple custom prompt with in-context examples, ensuring that the output only contains frame index ranges. An example instruction is as follows:
\begin{quote}
\texttt{The input images are frames from a video. Output the frame indexes that correspond to the text query: "{QUERY}". Only output the index range, for example, 2-4, 6-8.}
\end{quote}
We observe that GPT-4o follows this prompt reliably, typically producing clean and parseable frame ranges, which are then converted into time intervals for evaluation.

Unlike GPT-4o, Gemini models \cite{DBLP:journals/corr/abs-2312-11805} naturally support text, vision, and audio with extremely long context length, making them ideal for long video understanding. Gemini can also take raw video files as input. We evaluate two versions:
\begin{itemize}
    \item Gemini-2.5-Pro (0325): Offers strong reasoning and often outputs explicit Chain-of-Thought (CoT) explanations. However, when not constrained by output formatting instructions, the responses can be highly inconsistent and difficult to parse. In contrast, adding strict format requirements improves consistency but leads to degraded performance. Therefore, we choose the simple prompt without format requirement and parse the major possible formats for evaluation.
    \item Gemini-2.0-Flash (stable): Provides a better output structure and lower rejection rates, making it more reliable for batch evaluations. Although slightly less capable in reasoning than Gemini-2.5-Pro, it balances performance and robustness well.
\end{itemize}
The prompt provided to Gemini models is:
\begin{quote}
\texttt{Answer with time ranges and do not output explanation. What are all the time ranges corresponding to the text query: "{QUERY}"?}
\end{quote}
However, Gemini-0325 frequently deviates from these instructions, returning time ranges in inconsistent formats. We therefore implement careful parsing procedures to reliably extract time intervals for evaluation.

In Table \ref{tab:tr_results}, we report performance in multiple evaluation dimensions: video duration, query format, and query modality.
Remarkably, \method{} outperforms all baseline models in all categories for the primary metric $\bar{\text{IoU}}$. 
In ultra-short videos, Gemini-2.0-Flash achieves the best precision, but its recall is significantly lower, leading to a lower $\bar{\text{IoU}}$. Performance across different query formats is relatively consistent, though models generally struggle more with longer and more descriptive queries.
As expected, GPT-4o performs poorly in audio-only queries due to lack of audio input support. Other models perform comparably across modalities, but accuracy in audio-based queries is consistently lower than that on vision-based queries, highlighting the challenge of audio-conditioned understanding.

\begin{figure}[htbp]
  \centering
  \captionsetup[subfigure]{labelformat=empty}
  \begin{subfigure}{0.32\textwidth}
    \includegraphics[width=\linewidth]{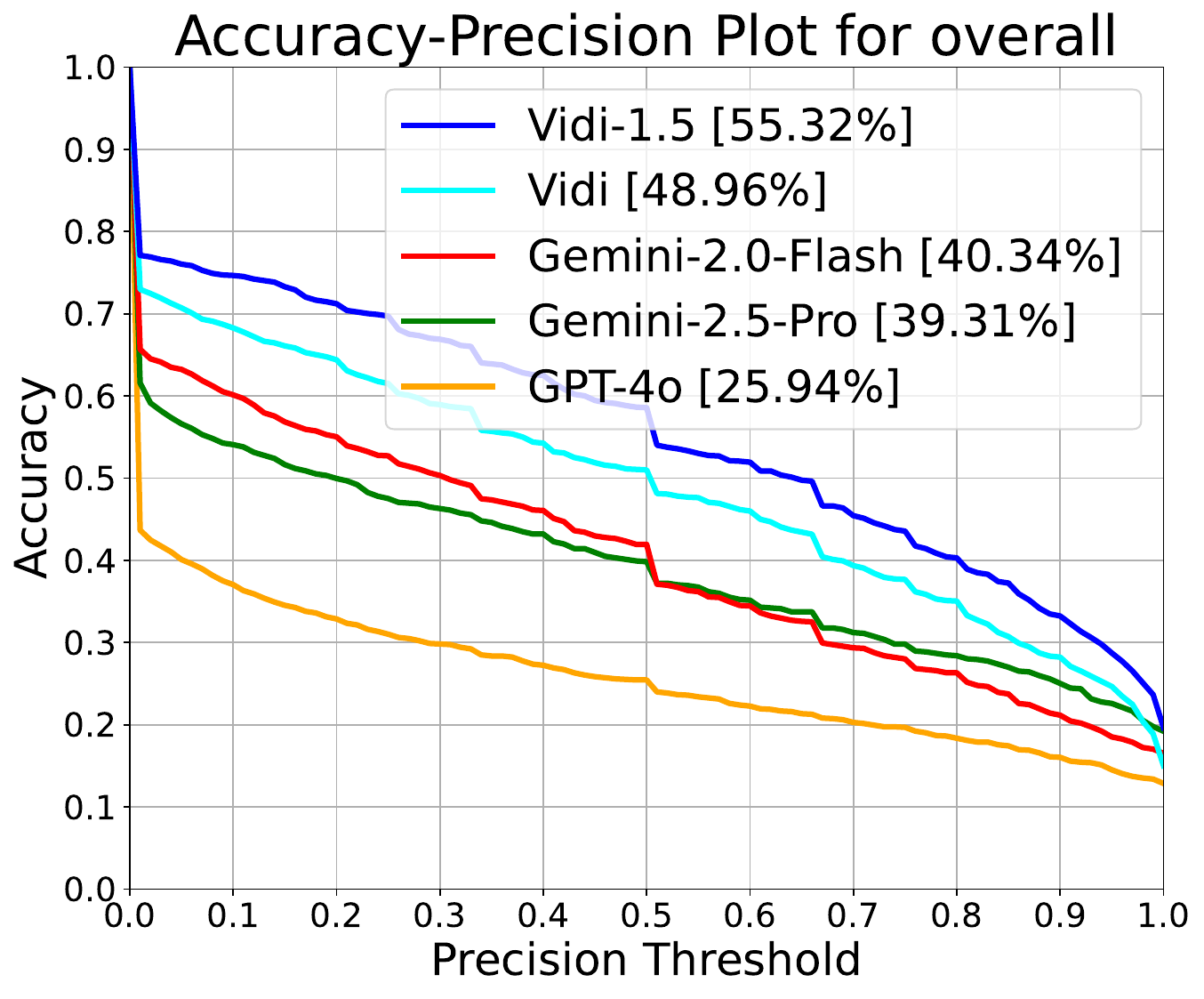}
    \label{fig:sub1}
  \end{subfigure}
  \hfill
  \begin{subfigure}{0.32\textwidth}
    \includegraphics[width=\linewidth]{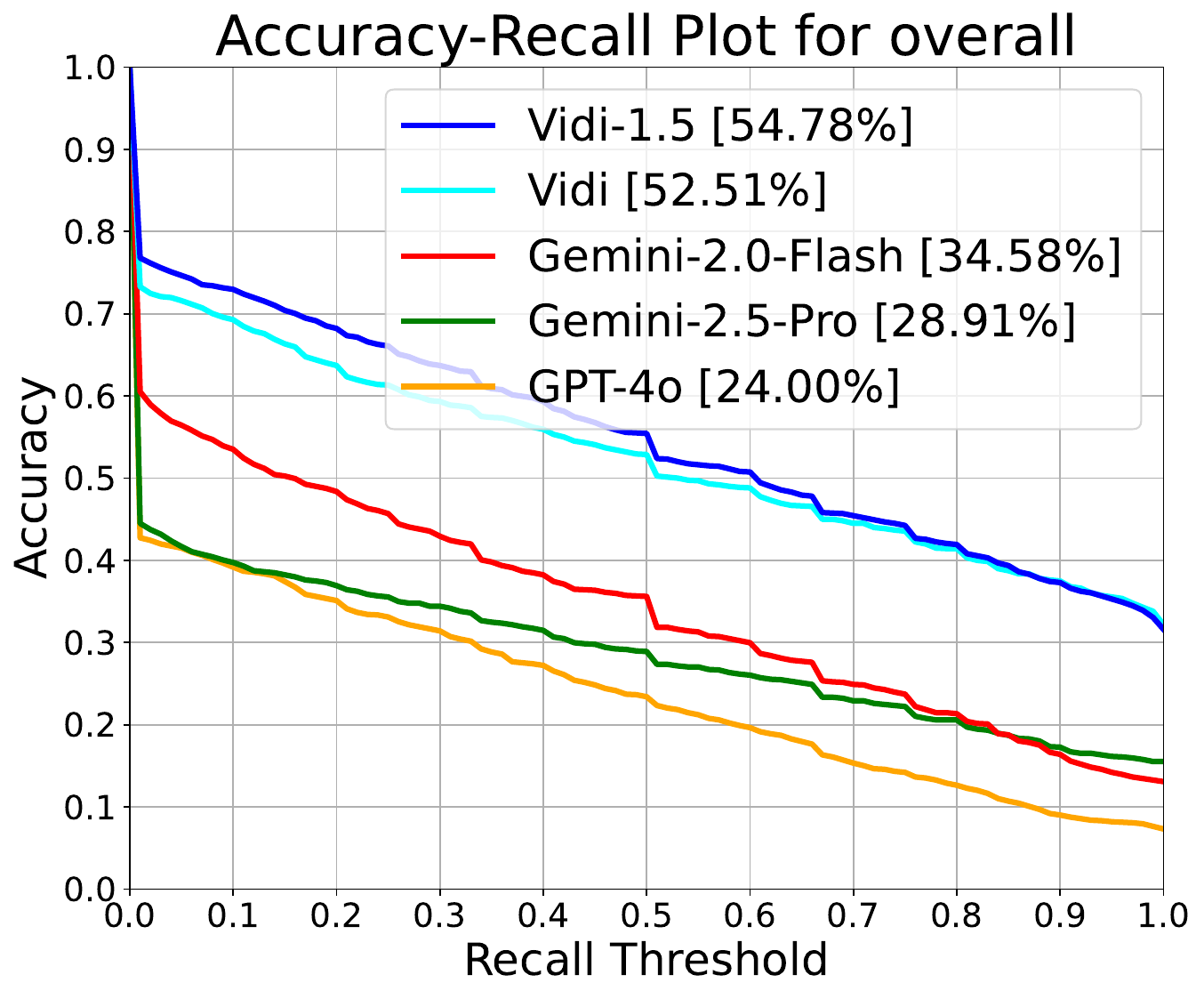}
    \label{fig:sub2}
  \end{subfigure}
  \hfill
  \begin{subfigure}{0.32\textwidth}
    \includegraphics[width=\linewidth]{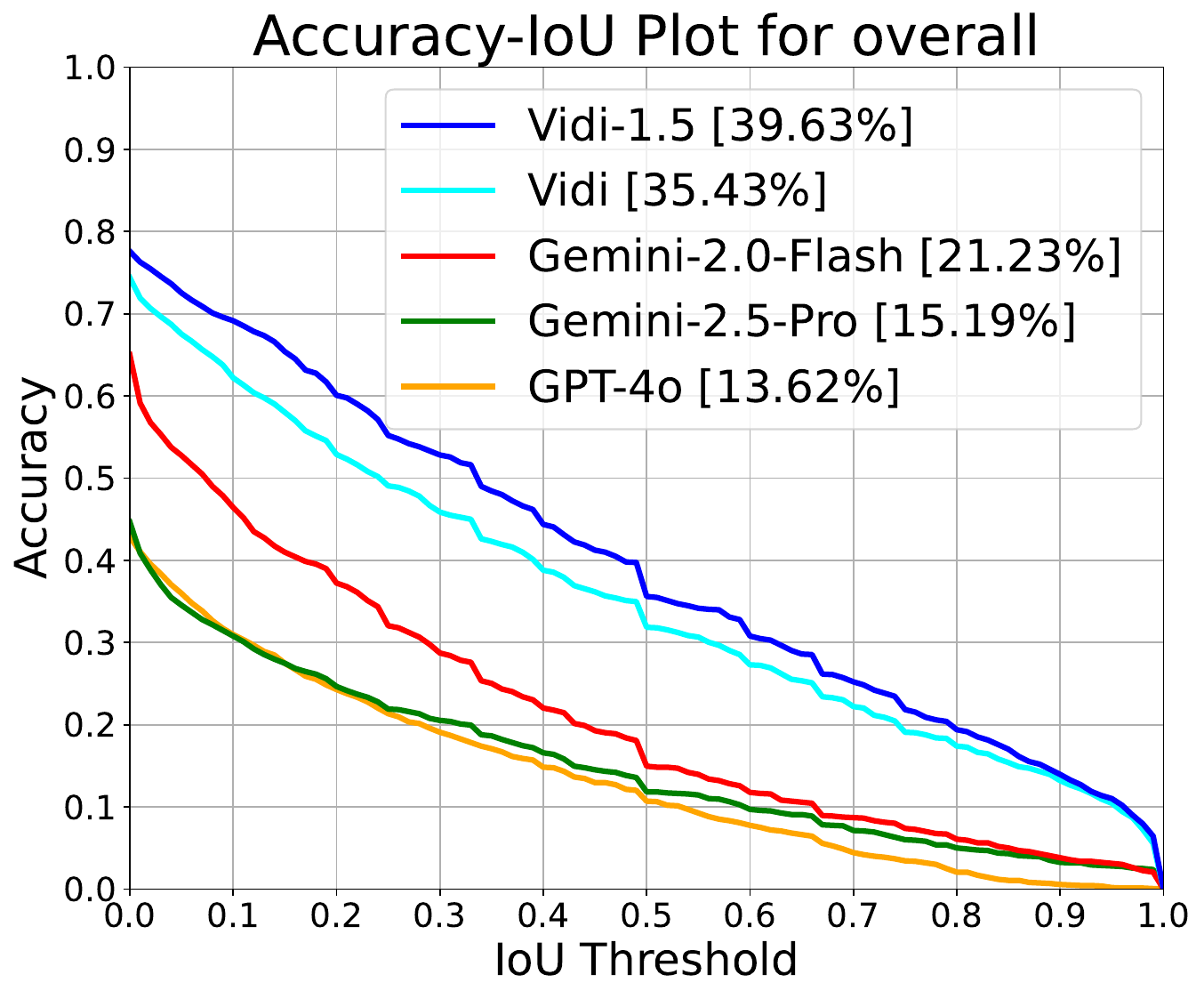}
    \label{fig:sub3}
  \end{subfigure}
  \vspace{-0.6cm}
  \caption{Overall performance curves for temporal retrieval on the VUE-TR benchmark. We report accuracy across varying thresholds for different models.}
  \label{fig:curves}
\end{figure}

As shown in Figure \ref{fig:curves}, we can still observe that \method{} consistently outperforms all competitors by a significant margin across the entire range of thresholds. This performance gap even widens at higher thresholds (\eg, $\geq 0.5$), where other models degrade much more rapidly. It shows \method{}'s strength in fine-grained temporal precision, a critical capability for real-world video editing and retrieval tasks. By using more powerful backbones and dynamic token compression, \method{}-1.5 gains considerable improvement over all the metrics.

%% file: sections/tables/hyperparameters.tex
\begin{table*}[!htbp]
\centering
\renewcommand{\arraystretch}{1.2}
\resizebox{0.9\linewidth}{!}{
\begin{tabular}{@{\extracolsep{4pt}}lccc@{}}
\toprule[1.5pt]
& \textbf{Adapter} & \textbf{Synthetic \& Real Videos} & \textbf{Post Training} \\
\midrule
lr adapter & 1e-3 & 5e-6 & 5e-6 \\
lr llm & frozen & 2e-6 & 2e-6  \\
mm encoders & frozen & frozen & frozen \\
video duration & 10-600 secs & 10-1800 secs & 10-1800 secs \\
weight decay & 0.0 & 0.0 & 0.0 \\
optimizer & AdamW \cite{DBLP:conf/iclr/LoshchilovH19} & AdamW \cite{DBLP:conf/iclr/LoshchilovH19} & AdamW \cite{DBLP:conf/iclr/LoshchilovH19} \\
optimizer $\beta_1$ & default (0.9) & default & default \\
optimizer $\beta_2$ & default (0.999) & default & default \\
optimizer $\epsilon$ & default (1e-8) & default & default \\
training steps & 2k & 10k & 1 epoch \\
lr scheduler & cosine & cosine & cosine \\
total batch size & 128 & 128 & 128 \\
dtype & bfloat16 & bfloat16 & bfloat16 \\
deepspeed & stage 2 & stage 3 & stage 3 \\
gradient ckpt & off & on & on \\
\bottomrule[1.5pt]
\end{tabular}
}
\caption{
Training configurations and hyper-parameters used across different training stages for \method{}.
}\label{tab:hyperparams}
\end{table*}

%% file: sections/related_work.tex
\section{Related Work}
{\flushleft \textbf{Benchmarks.}}
Recent advancements in video question answering (VideoQA) have been sharped by benchmarks such as CinePile \cite{DBLP:journals/corr/abs-2405-08813}, Video-MME \cite{DBLP:journals/corr/abs-2405-21075}, MovieChat-1K \cite{DBLP:conf/cvpr/SongCWZZWCG0ZLH24}, LVBench \cite{DBLP:journals/corr/abs-2406-08035}, and LongVideoBench \cite{DBLP:conf/nips/WuLCL24}.
To evaluate models on long-form videos, many of these benchmarks demonstrate strong performance with only sparse frame sampling. They may not fully reflect a model's video understanding capabilities. For example, LLaVA-OneVision \cite{DBLP:journals/corr/abs-2408-03326} only employs as few as $32$ uniformly sampled frames to achieve $66.3\%$ accuracy on Video-MME \cite{DBLP:journals/corr/abs-2405-21075}.

In contrast, temporal retrieval requires fine-grained understanding across the full temporal span of a video instead of a few scattered glimpses. As such, temporal retrieval serves as a more rigorous benchmark for evaluating a model's true capability in video understanding.
Charades-STA \cite{charades} and DiDeMo \cite{DBLP:conf/iccv/HendricksWSSDR17} focus mainly on action grounding within short clips ($\sim30s$). QVHighlights \cite{qvhighlight} and ActivityNet Captions \cite{DBLP:conf/iccv/KrishnaHRFN17} extend to several minutes videos, covering event descriptions and subjectively defined highlights. Recently, LongVALE \cite{DBLP:journals/corr/abs-2411-19772} introduces multimodal audio-visual queries, but remains constrained to relatively short durations.

To address above limitations, our benchmark is designed to evaluate model performance in realistic, large-scale video environments. 
In particular, we extend the video length to over one hour, and introduce a video duration categorization into ultra-short, short, medium, long, and ultra-long segments. 
Furthermore, we diversify the retrieval challenge by varying query format (keyword, phrase, sentence) and modality (vision-only, audio-only, vision+audio), enabling a more comprehensive understanding of model capabilities. 

{\flushleft \textbf{Video LMMs.}}
Video understanding has made significant progress with the emergence of LMMs that integrate visual and textual information. Models such as the InternLM-XComposer \cite{DBLP:journals/corr/abs-2309-15112, DBLP:journals/corr/abs-2401-16420, zhang2023internlm}, Qwen-VL \cite{DBLP:journals/corr/abs-2308-12966, DBLP:journals/corr/abs-2409-12191, bai2025qwen2}, and LLaVA \cite{DBLP:conf/nips/LiuLWL23a, DBLP:conf/nips/WuLCL24} series have introduced innovative architectures and training strategies.
InternLM-XComposer adopts a modular architecture that integrates interleaved text-image composition and comprehension with rich multilingual knowledge.
Qwen-VL and its successors introduce flexible multi-image and box-conditioned input capabilities, supporting multilingual and multi-turn interaction for diverse vision-language tasks.
The LLaVA series bridges visual perception with LLMs through a unified representation space, and its recent extensions, such as LLaVA-NeXT, further support video and 3D input via interleaved multimodal pretraining \cite{DBLP:journals/corr/abs-2407-07895}.
Although these models demonstrate impressive general-purpose reasoning, they typically operate on a limited number of frames (\eg, $\leq64$), which restricts their effectiveness in long-form temporal retrieval.

{\flushleft \textbf{Temporal Retrieval.}} 
Early methods \cite{qvhighlight, charades, DBLP:conf/iccv/HendricksWSSDR17, DBLP:conf/iccv/KrishnaHRFN17} usually employ proposal-based strategies, where models first generate candidate segments and then rank them based on their semantic relevance to the query.
With the emergence of video LMMs, recent research has shifted toward enhancing video-text alignment through unified frameworks. For example, TimeChat~\cite{ren2024timechat} connects a time-aware frame encoder and a sliding video Q-Former \cite{DBLP:conf/icml/0008LSH23} to deal with temporal retrieval based on an instruction-tuning dataset.
TimeSearch~\cite{pan2025timesearch} proposes a temporal spotlight grounding strategy to find key events and a temporal reflection mechanism to verify time range predictions and guide the search direction.
Ye~\etal~\cite{ye2025re} revisit the temporal search in long-form videos by proposing a lightweight framework that reformulates temporal retrieval as a spatial search problem. 
Although efficient, the above methods rely on sampling only $8\sim96$ frames, which may be insufficient to capture fine-grained temporal cues.

Despite these advancements, most existing methods still focus on short videos or rely on sparse frame sampling, and primarily limited to vision-only inputs. These constraints hinder the evaluation of models' full multimodal reasoning abilities, particularly in understanding and retrieving moments from long, complex videos involving both visual and auditory modalities.
In this work, we propose \method{}, a new approach that jointly incorporates text, vision, and audio input from hour-long videos to retrieve the most relevant temporal segments based on free-form natural language queries. Our setting pushes beyond existing benchmarks by requiring fine-grained temporal understanding across extended multimodal content.

%% file: sections/conclusion.tex
\section{Conclusion}
We introduce \method{}, a large multi-modal model for real-world video understanding and editing (VUE). The first release focuses on temporal retrieval (TR), which is a foundational step in video trimming and editing, by localizing relevant segments from long videos given natural language queries.
To support robust evaluation under realistic conditions, we propose VUE-TR, a new benchmark for practical video editing use cases. VUE-TR introduces key improvements over prior datasets in terms of video duration, audio support, query format, annotation quality, and metric design for multispan timestamp evaluation. 
\method{} leverages a unified vision-audio-language architecture with modality-aware sampling and precise token alignment. In addition, the Decomposed Attention mechanism allows the model to capture fine-grained temporal cues across diverse modalities and scales seamlessly to long videos. 
As demonstrated in the VUE-TR benchmark, \method{} consistently outperforms leading proprietary models such as GPT-4o and Gemini.
For future work, we plan to extend \method{} to support a wider spectrum of video understanding tasks, including temporal VQA, spatio-temporal grounding, and high-level video understanding. Additionally, we aim to explore interactive editing capabilities to further bridge the gap between large models and real-world VUE applications.

%% file: sections/contributors.tex
\clearpage
\section{Contributors}
\label{sec:contributor}
\textbf{Core Contributors - Research (alphabetical order)}\\
Chia-Wen Kuo, Dawei Du, Fan Chen, Guang Chen, Sijie Zhu, Xin Gu. \\

\noindent\textbf{Core Contributors - Infrastructure (alphabetical order)} \\
Celong Liu, Tong Jin. \\

\noindent\textbf{Research Leads} \\
Longyin Wen, Xiaohui Shen.\\

\noindent\textbf{Contributors (alphabetical order)}\\
Jiamin Yuan, Lingxi Zhang, Lu Guo, Lusha Li, Qingyu Chen, Rachel Deng, Stuart Siew, Wei Lu, Wen Zhong, Xing Mei, Xueqiong Qu, Zhenfang Chen.

%% file: main.bbl
\begin{thebibliography}{10}

\bibitem{DBLP:journals/corr/abs-2312-11805}
Rohan Anil, Sebastian Borgeaud, Yonghui Wu, Jean{-}Baptiste Alayrac, Jiahui Yu, Radu Soricut, Johan Schalkwyk, Andrew~M. Dai, Anja Hauth, Katie Millican, David Silver, Slav Petrov, Melvin Johnson, Ioannis Antonoglou, Julian Schrittwieser, Amelia Glaese, Jilin Chen, Emily Pitler, Timothy~P. Lillicrap, Angeliki Lazaridou, Orhan Firat, James Molloy, Michael Isard, Paul~Ronald Barham, Tom Hennigan, Benjamin Lee, Fabio Viola, Malcolm Reynolds, Yuanzhong Xu, Ryan Doherty, Eli Collins, Clemens Meyer, Eliza Rutherford, Erica Moreira, Kareem Ayoub, Megha Goel, George Tucker, Enrique Piqueras, Maxim Krikun, Iain Barr, Nikolay Savinov, Ivo Danihelka, Becca Roelofs, Ana{\"{\i}}s White, Anders Andreassen, Tamara von Glehn, Lakshman Yagati, Mehran Kazemi, Lucas Gonzalez, Misha Khalman, Jakub Sygnowski, and et~al.
\newblock Gemini: {A} family of highly capable multimodal models.
\newblock {\em CoRR}, abs/2312.11805, 2023.

\bibitem{DBLP:journals/corr/abs-2308-12966}
Jinze Bai, Shuai Bai, Shusheng Yang, Shijie Wang, Sinan Tan, Peng Wang, Junyang Lin, Chang Zhou, and Jingren Zhou.
\newblock Qwen-vl: {A} frontier large vision-language model with versatile abilities.
\newblock {\em CoRR}, abs/2308.12966, 2023.

\bibitem{bai2025qwen2}
Shuai Bai, Keqin Chen, Xuejing Liu, Jialin Wang, Wenbin Ge, Sibo Song, Kai Dang, Peng Wang, Shijie Wang, Jun Tang, Humen Zhong, Yuanzhi Zhu, Ming{-}Hsuan Yang, Zhaohai Li, Jianqiang Wan, Pengfei Wang, Wei Ding, Zheren Fu, Yiheng Xu, Jiabo Ye, Xi~Zhang, Tianbao Xie, Zesen Cheng, Hang Zhang, Zhibo Yang, Haiyang Xu, and Junyang Lin.
\newblock Qwen2.5-vl technical report.
\newblock {\em CoRR}, abs/2502.13923, 2025.

\bibitem{DBLP:journals/corr/abs-2401-16420}
Xiaoyi Dong, Pan Zhang, Yuhang Zang, Yuhang Cao, Bin Wang, Linke Ouyang, Xilin Wei, Songyang Zhang, Haodong Duan, Maosong Cao, Wenwei Zhang, Yining Li, Hang Yan, Yang Gao, Xinyue Zhang, Wei Li, Jingwen Li, Kai Chen, Conghui He, Xingcheng Zhang, Yu~Qiao, Dahua Lin, and Jiaqi Wang.
\newblock Internlm-xcomposer2: Mastering free-form text-image composition and comprehension in vision-language large model.
\newblock {\em CoRR}, abs/2401.16420, 2024.

\bibitem{DBLP:journals/corr/abs-2405-21075}
Chaoyou Fu, Yuhan Dai, Yondong Luo, Lei Li, Shuhuai Ren, Renrui Zhang, Zihan Wang, Chenyu Zhou, Yunhang Shen, Mengdan Zhang, Peixian Chen, Yanwei Li, Shaohui Lin, Sirui Zhao, Ke~Li, Tong Xu, Xiawu Zheng, Enhong Chen, Rongrong Ji, and Xing Sun.
\newblock Video-mme: The first-ever comprehensive evaluation benchmark of multi-modal llms in video analysis.
\newblock {\em CoRR}, abs/2405.21075, 2024.

\bibitem{charades}
Jiyang Gao, Chen Sun, Zhenheng Yang, and Ram Nevatia.
\newblock {TALL:} temporal activity localization via language query.
\newblock In {\em {IEEE} International Conference on Computer Vision, {ICCV} 2017, Venice, Italy, October 22-29, 2017}, pages 5277--5285. {IEEE} Computer Society, 2017.

\bibitem{DBLP:journals/corr/abs-2411-19772}
Tiantian Geng, Jinrui Zhang, Qingni Wang, Teng Wang, Jinming Duan, and Feng Zheng.
\newblock Longvale: Vision-audio-language-event benchmark towards time-aware omni-modal perception of long videos.
\newblock {\em CoRR}, abs/2411.19772, 2024.

\bibitem{guhr-EtAl:2021:fullstop}
Oliver Guhr, Anne{-}Kathrin Schumann, Frank Bahrmann, and Hans{-}Joachim B{\"{o}}hme.
\newblock Fullstop: Multilingual deep models for punctuation prediction.
\newblock In {\em Proceedings of the Swiss Text Analytics Conference 2021, Winterthur, Switzerland, June 14-16, 2021 (held online due to {COVID19} pandemic)}, volume 2957 of {\em {CEUR} Workshop Proceedings}. CEUR-WS.org, 2021.

\bibitem{DBLP:conf/iccv/HendricksWSSDR17}
Lisa~Anne Hendricks, Oliver Wang, Eli Shechtman, Josef Sivic, Trevor Darrell, and Bryan~C. Russell.
\newblock Localizing moments in video with natural language.
\newblock In {\em {IEEE} International Conference on Computer Vision, {ICCV} 2017, Venice, Italy, October 22-29, 2017}, pages 5804--5813. {IEEE} Computer Society, 2017.

\bibitem{DBLP:journals/corr/abs-2410-21276}
Aaron Hurst, Adam Lerer, Adam~P. Goucher, Adam Perelman, Aditya Ramesh, Aidan Clark, AJ~Ostrow, Akila Welihinda, Alan Hayes, Alec Radford, Aleksander Madry, Alex Baker{-}Whitcomb, and et~al.
\newblock Gpt-4o system card.
\newblock {\em CoRR}, abs/2410.21276, 2024.

\bibitem{jiang2023mistral}
Albert~Q. Jiang, Alexandre Sablayrolles, Arthur Mensch, Chris Bamford, Devendra~Singh Chaplot, Diego de~Las~Casas, Florian Bressand, Gianna Lengyel, Guillaume Lample, Lucile Saulnier, L{\'{e}}lio~Renard Lavaud, Marie{-}Anne Lachaux, Pierre Stock, Teven~Le Scao, Thibaut Lavril, Thomas Wang, Timoth{\'{e}}e Lacroix, and William~El Sayed.
\newblock Mistral 7b.
\newblock {\em CoRR}, abs/2310.06825, 2023.

\bibitem{DBLP:conf/iccv/KrishnaHRFN17}
Ranjay Krishna, Kenji Hata, Frederic Ren, Li~Fei{-}Fei, and Juan~Carlos Niebles.
\newblock Dense-captioning events in videos.
\newblock In {\em {IEEE} International Conference on Computer Vision, {ICCV} 2017, Venice, Italy, October 22-29, 2017}, pages 706--715. {IEEE} Computer Society, 2017.

\bibitem{kuo2025rethinking}
Chia{-}Wen Kuo, Sijie Zhu, Fan Chen, Xiaohui Shen, and Longyin Wen.
\newblock Rethinking homogeneity of vision and text tokens in large vision-and-language models.
\newblock {\em CoRR}, abs/2502.01906, 2025.

\bibitem{qvhighlight}
Jie Lei, Tamara~L. Berg, and Mohit Bansal.
\newblock Detecting moments and highlights in videos via natural language queries.
\newblock In {\em Advances in Neural Information Processing Systems 34: Annual Conference on Neural Information Processing Systems 2021, NeurIPS 2021, December 6-14, 2021, virtual}, pages 11846--11858, 2021.

\bibitem{DBLP:journals/corr/abs-2408-03326}
Bo~Li, Yuanhan Zhang, Dong Guo, Renrui Zhang, Feng Li, Hao Zhang, Kaichen Zhang, Yanwei Li, Ziwei Liu, and Chunyuan Li.
\newblock Llava-onevision: Easy visual task transfer.
\newblock {\em CoRR}, abs/2408.03326, 2024.

\bibitem{li2022structured}
Congcong Li, Xinyao Wang, Dexiang Hong, Yufei Wang, Libo Zhang, Tiejian Luo, and Longyin Wen.
\newblock Structured context transformer for generic event boundary detection.
\newblock {\em CoRR}, abs/2206.02985, 2022.

\bibitem{DBLP:journals/corr/abs-2407-07895}
Feng Li, Renrui Zhang, Hao Zhang, Yuanhan Zhang, Bo~Li, Wei Li, Zejun Ma, and Chunyuan Li.
\newblock Llava-next-interleave: Tackling multi-image, video, and 3d in large multimodal models.
\newblock {\em CoRR}, abs/2407.07895, 2024.

\bibitem{DBLP:conf/icml/0008LSH23}
Junnan Li, Dongxu Li, Silvio Savarese, and Steven C.~H. Hoi.
\newblock {BLIP-2:} bootstrapping language-image pre-training with frozen image encoders and large language models.
\newblock In {\em International Conference on Machine Learning, {ICML} 2023, 23-29 July 2023, Honolulu, Hawaii, {USA}}, volume 202 of {\em Proceedings of Machine Learning Research}, pages 19730--19742. {PMLR}, 2023.

\bibitem{DBLP:conf/nips/LiuLWL23a}
Haotian Liu, Chunyuan Li, Qingyang Wu, and Yong~Jae Lee.
\newblock Visual instruction tuning.
\newblock In {\em Advances in Neural Information Processing Systems 36: Annual Conference on Neural Information Processing Systems 2023, NeurIPS 2023, New Orleans, LA, USA, December 10 - 16, 2023}, 2023.

\bibitem{DBLP:conf/iclr/LoshchilovH19}
Ilya Loshchilov and Frank Hutter.
\newblock Decoupled weight decay regularization.
\newblock In {\em 7th International Conference on Learning Representations, {ICLR} 2019, New Orleans, LA, USA, May 6-9, 2019}. OpenReview.net, 2019.

\bibitem{pan2025timesearch}
Junwen Pan, Rui Zhang, Xin Wan, Yuan Zhang, Ming Lu, and Qi~She.
\newblock Timesearch: Hierarchical video search with spotlight and reflection for human-like long video understanding.
\newblock {\em CoRR}, abs/2504.01407, 2025.

\bibitem{radford2023robust}
Alec Radford, Jong~Wook Kim, Tao Xu, Greg Brockman, Christine McLeavey, and Ilya Sutskever.
\newblock Robust speech recognition via large-scale weak supervision.
\newblock In {\em International Conference on Machine Learning, {ICML} 2023, 23-29 July 2023, Honolulu, Hawaii, {USA}}, volume 202 of {\em Proceedings of Machine Learning Research}, pages 28492--28518. {PMLR}, 2023.

\bibitem{DBLP:journals/corr/abs-2405-08813}
Ruchit Rawal, Khalid Saifullah, Ronen Basri, David Jacobs, Gowthami Somepalli, and Tom Goldstein.
\newblock Cinepile: {A} long video question answering dataset and benchmark.
\newblock {\em CoRR}, abs/2405.08813, 2024.

\bibitem{ren2024timechat}
Shuhuai Ren, Linli Yao, Shicheng Li, Xu~Sun, and Lu~Hou.
\newblock Timechat: {A} time-sensitive multimodal large language model for long video understanding.
\newblock In {\em {IEEE/CVF} Conference on Computer Vision and Pattern Recognition, {CVPR} 2024, Seattle, WA, USA, June 16-22, 2024}, pages 14313--14323. {IEEE}, 2024.

\bibitem{team2024gemma}
Morgane Rivi{\`{e}}re, Shreya Pathak, Pier~Giuseppe Sessa, Cassidy Hardin, Surya Bhupatiraju, L{\'{e}}onard Hussenot, Thomas Mesnard, Bobak Shahriari, Alexandre Ram{\'{e}}, and et~al.
\newblock Gemma 2: Improving open language models at a practical size.
\newblock {\em CoRR}, abs/2408.00118, 2024.

\bibitem{DBLP:conf/cvpr/SongCWZZWCG0ZLH24}
Enxin Song, Wenhao Chai, Guanhong Wang, Yucheng Zhang, Haoyang Zhou, Feiyang Wu, Haozhe Chi, Xun Guo, Tian Ye, Yanting Zhang, Yan Lu, Jenq{-}Neng Hwang, and Gaoang Wang.
\newblock Moviechat: From dense token to sparse memory for long video understanding.
\newblock In {\em {IEEE/CVF} Conference on Computer Vision and Pattern Recognition, {CVPR} 2024, Seattle, WA, USA, June 16-22, 2024}, pages 18221--18232. {IEEE}, 2024.

\bibitem{DBLP:journals/corr/abs-2502-14786}
Michael Tschannen, Alexey~A. Gritsenko, Xiao Wang, Muhammad~Ferjad Naeem, Ibrahim Alabdulmohsin, Nikhil Parthasarathy, Talfan Evans, Lucas Beyer, Ye~Xia, Basil Mustafa, Olivier~J. H{\'{e}}naff, Jeremiah Harmsen, Andreas Steiner, and Xiaohua Zhai.
\newblock Siglip 2: Multilingual vision-language encoders with improved semantic understanding, localization, and dense features.
\newblock {\em CoRR}, abs/2502.14786, 2025.

\bibitem{DBLP:conf/nips/VaswaniSPUJGKP17}
Ashish Vaswani, Noam Shazeer, Niki Parmar, Jakob Uszkoreit, Llion Jones, Aidan~N. Gomez, Lukasz Kaiser, and Illia Polosukhin.
\newblock Attention is all you need.
\newblock In {\em Advances in Neural Information Processing Systems 30: Annual Conference on Neural Information Processing Systems 2017, December 4-9, 2017, Long Beach, CA, {USA}}, pages 5998--6008, 2017.

\bibitem{DBLP:journals/corr/abs-2409-12191}
Peng Wang, Shuai Bai, Sinan Tan, Shijie Wang, Zhihao Fan, Jinze Bai, Keqin Chen, Xuejing Liu, Jialin Wang, Wenbin Ge, Yang Fan, Kai Dang, Mengfei Du, Xuancheng Ren, Rui Men, Dayiheng Liu, Chang Zhou, Jingren Zhou, and Junyang Lin.
\newblock Qwen2-vl: Enhancing vision-language model's perception of the world at any resolution.
\newblock {\em CoRR}, abs/2409.12191, 2024.

\bibitem{DBLP:journals/corr/abs-2410-08260}
Qiuheng Wang, Yukai Shi, Jiarong Ou, Rui Chen, Ke~Lin, Jiahao Wang, Boyuan Jiang, Haotian Yang, Mingwu Zheng, Xin Tao, Fei Yang, Pengfei Wan, and Di~Zhang.
\newblock Koala-36m: {A} large-scale video dataset improving consistency between fine-grained conditions and video content.
\newblock {\em CoRR}, abs/2410.08260, 2024.

\bibitem{DBLP:journals/corr/abs-2406-08035}
Weihan Wang, Zehai He, Wenyi Hong, Yean Cheng, Xiaohan Zhang, Ji~Qi, Shiyu Huang, Bin Xu, Yuxiao Dong, Ming Ding, and Jie Tang.
\newblock Lvbench: An extreme long video understanding benchmark.
\newblock {\em CoRR}, abs/2406.08035, 2024.

\bibitem{DBLP:conf/nips/Wei0SBIXCLZ22}
Jason Wei, Xuezhi Wang, Dale Schuurmans, Maarten Bosma, Brian Ichter, Fei Xia, Ed~H. Chi, Quoc~V. Le, and Denny Zhou.
\newblock Chain-of-thought prompting elicits reasoning in large language models.
\newblock In {\em Advances in Neural Information Processing Systems 35: Annual Conference on Neural Information Processing Systems 2022, NeurIPS 2022, New Orleans, LA, USA, November 28 - December 9, 2022}, 2022.

\bibitem{DBLP:conf/nips/WuLCL24}
Haoning Wu, Dongxu Li, Bei Chen, and Junnan Li.
\newblock Longvideobench: {A} benchmark for long-context interleaved video-language understanding.
\newblock In {\em Advances in Neural Information Processing Systems 38: Annual Conference on Neural Information Processing Systems 2024, NeurIPS 2024, Vancouver, BC, Canada, December 10 - 15, 2024}, 2024.

\bibitem{ye2025re}
Jinhui Ye, Zihan Wang, Haosen Sun, Keshigeyan Chandrasegaran, Zane Durante, Cristobal Eyzaguirre, Yonatan Bisk, Juan~Carlos Niebles, Ehsan Adeli, Li~Fei-Fei, Jiajun Wu, and Manling Li.
\newblock Re-thinking temporal search for long-form video understanding.
\newblock {\em CoRR}, abs/2504.02259, 2025.

\bibitem{zhai2023sigmoid}
Xiaohua Zhai, Basil Mustafa, Alexander Kolesnikov, and Lucas Beyer.
\newblock Sigmoid loss for language image pre-training.
\newblock In {\em {IEEE/CVF} International Conference on Computer Vision, {ICCV} 2023, Paris, France, October 1-6, 2023}, pages 11941--11952. {IEEE}, 2023.

\bibitem{DBLP:journals/corr/abs-2309-15112}
Pan Zhang, Xiaoyi Dong, Bin Wang, Yuhang Cao, Chao Xu, Linke Ouyang, Zhiyuan Zhao, Shuangrui Ding, Songyang Zhang, Haodong Duan, Wenwei Zhang, Hang Yan, Xinyue Zhang, Wei Li, Jingwen Li, Kai Chen, Conghui He, Xingcheng Zhang, Yu~Qiao, Dahua Lin, and Jiaqi Wang.
\newblock Internlm-xcomposer: {A} vision-language large model for advanced text-image comprehension and composition.
\newblock {\em CoRR}, abs/2309.15112, 2023.

\bibitem{zhang2023internlm}
Pan Zhang, Xiaoyi Dong, Yuhang Zang, Yuhang Cao, Rui Qian, Lin Chen, Qipeng Guo, Haodong Duan, Bin Wang, Linke Ouyang, Songyang Zhang, Wenwei Zhang, Yining Li, Yang Gao, Peng Sun, Xinyue Zhang, Wei Li, Jingwen Li, Wenhai Wang, Hang Yan, Conghui He, Xingcheng Zhang, Kai Chen, Jifeng Dai, Yu~Qiao, Dahua Lin, and Jiaqi Wang.
\newblock Internlm-xcomposer-2.5: {A} versatile large vision language model supporting long-contextual input and output.
\newblock {\em CoRR}, abs/2407.03320, 2024.

\end{thebibliography}
